\def\archiveversion{}
\documentclass{article}

\usepackage{iclr2027_conference,times}
\usepackage[T1]{fontenc}
\usepackage[utf8]{inputenc}
\usepackage{microtype}
\IfFileExists{inconsolata.sty}{\usepackage{inconsolata}}{}
\usepackage{graphicx}
\usepackage{url}
\usepackage{booktabs}
\usepackage{amsfonts}
\usepackage{amsmath}
\usepackage{amssymb}
\usepackage{amsthm}
\usepackage{mathtools}
\usepackage{xcolor}
\usepackage[nointegrals]{wasysym}   
\usepackage{float}
\usepackage{placeins}
\usepackage{subcaption}
\usepackage{tikz}
\usetikzlibrary{arrows.meta,fit,positioning}
\usepackage{enumitem}
\usepackage{algorithm}
\usepackage{algorithmic}
\usepackage{listings}
\usepackage{hyperref}
\hypersetup{hidelinks}
\usepackage{cleveref}
\usepackage{tabularx,booktabs}

\IfFileExists{wrapfig.sty}{%
  \usepackage{wrapfig}%
}{%
  \newenvironment{wrapfigure}[2]{\begin{figure}[t]\centering}{\end{figure}}%
  \newenvironment{wraptable}[2]{\begin{table}[t]\centering}{\end{table}}%
}

\newcommand{\Cov}{\operatorname{Cov}}

\newcommand{\voirdire}{\textup{VoirDire}}

\newtheorem{theorem}{Theorem}
\newtheorem{proposition}{Proposition}

\newtheorem{corollary}{Corollary}
\newcommand{\fullstatementtitle}{}
\newtheorem*{fullstatementbody}{\fullstatementtitle}
\newenvironment{fullstatement}[1]
  {\renewcommand{\fullstatementtitle}{#1}\begin{samepage}\begin{fullstatementbody}[Full statement]}
  {\end{fullstatementbody}\end{samepage}}
\crefname{algorithm}{algorithm}{algorithms}

\lstdefinestyle{prompt}{
  basicstyle=\footnotesize\ttfamily,
  breaklines=true,
  columns=fullflexible,
  keepspaces=true,
  showstringspaces=false,
  frame=none,
  xleftmargin=0.5em,
  aboveskip=0.35\baselineskip,
  belowskip=0.35\baselineskip
}
\crefname{section}{Section}{Sections}
\Crefname{section}{Section}{Sections}
\crefname{proposition}{Proposition}{Propositions}
\crefname{theorem}{Theorem}{Theorems}
\crefname{corollary}{Corollary}{Corollaries}
\crefname{lemma}{Lemma}{Lemmas}
\crefname{appendix}{Appendix}{Appendices}
\Crefname{appendix}{Appendix}{Appendices}

\title{\voirdire{}: Counterfactual Calibration for\\
Reliable LLM-as-a-Judge Aggregation}

\ifdefined\archiveversion
\author{Changho Shin$^{1}$ \quad Frederic Sala$^{2}$ \\[4pt]
\footnotesize\normalfont $^{1}$Department of Computer Science, Princeton University, Princeton, NJ \\
\footnotesize\normalfont $^{2}$Department of Computer Sciences, University of Wisconsin--Madison, Madison, WI}
\else
\author{Anonymous Authors}
\fi

\ifdefined\archiveversion\iclrfinalcopy\fi

\begin{document}
\maketitle
\ifdefined\archiveversion\lhead{}\fi


\begin{abstract}
LLM-as-a-judge systems often aggregate multiple judges to improve evaluation reliability.
However, judge agreement can reflect shared biases toward verbosity, formatting, or fluency rather than the intended evaluation target.
Addressing these shared biases is challenging because observational judge outputs alone cannot distinguish agreement driven by target quality from agreement driven by nuisance attributes.
We introduce \voirdire{}, an intervention-calibrated aggregation method for LLM-judge panels.
\voirdire{} applies controlled interventions that vary specified surface attributes while aiming to preserve the evaluation target; it then uses paired changes in judge scores to identify nuisance-sensitive directions.
It attenuates these directions and selects aggregation weights that retain variation across items while penalizing sensitivity to the interventions, making the final consensus less dependent on superficial features.
We show theoretically that observational judge outputs alone cannot identify which shared direction is target-relevant, while paired interventions provide the contrast needed to identify nuisance-sensitive variation.
We evaluate \voirdire{} on 12 datasets spanning continuous scoring, binary classification, and pairwise preference.
Compared with the existing LLM-judge aggregation baselines, \voirdire{} reduces MAE by up to 26.8\% on continuous-score tasks and improves accuracy by up to 5.3 percentage points on binary and preference tasks.
These findings suggest that counterfactual interventions provide an additional means of improving the reliability and controllability of LLM-based evaluation beyond merely relying on observed judge agreement.
\end{abstract}

\section{Introduction}
\begin{wrapfigure}{r}{0.5\textwidth}
    \vspace{-1.2em}
    \centering
    \includegraphics[width=0.48\textwidth]{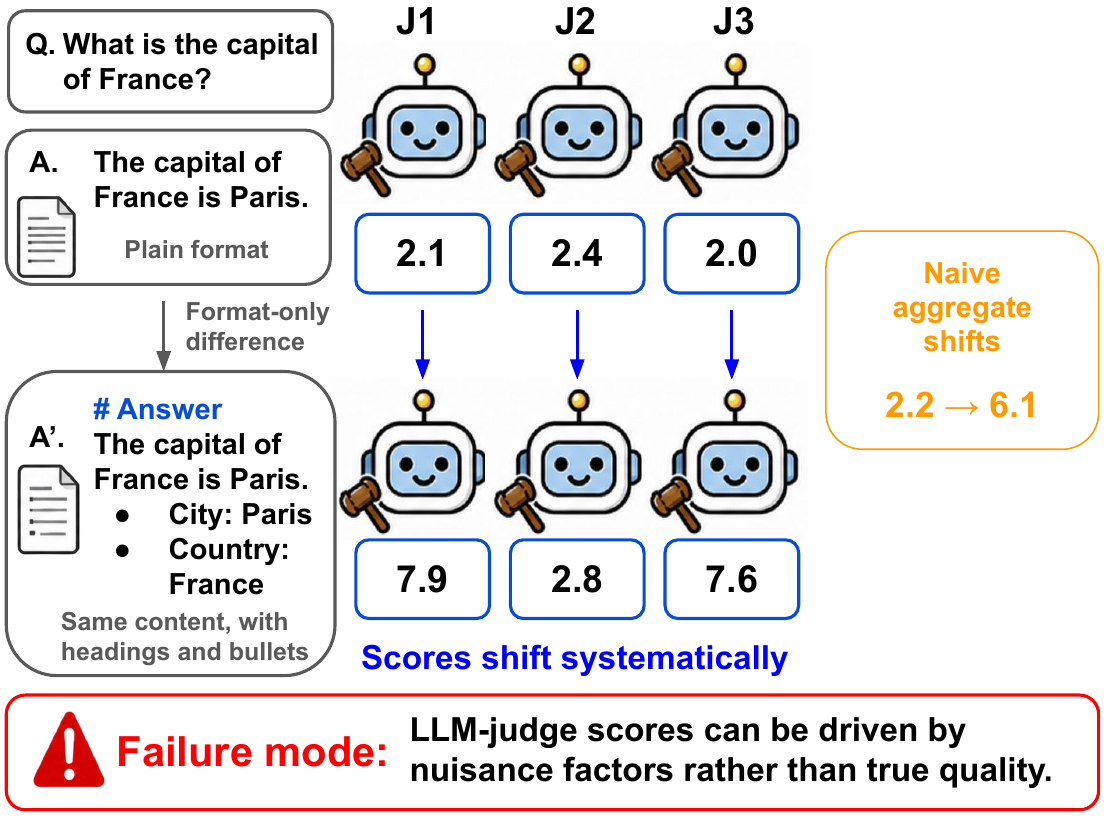}
    \vspace{-0.6em}
    \caption{\textbf{Motivation.} Surface attribute changes systematically shift judge scores, so aggregation can track nuisance factors rather than the evaluation target.}
    \label{fig:motivation}
    \vspace{-0.8em}
\end{wrapfigure}
LLM-as-a-judge systems are increasingly used to score open-ended answers, compare assistants, filter data, and approximate human evaluation \citep{liu2023geval,zheng2023judging,wang2024pandalm,zhu2025judgelm}. Because a single judge can be noisy, recent work has explored aggregating multiple LLM judges \citep{verga2024replacing,li2024prd,ai2025beyond,balasubramanian2026dependence,zhao2026care}. The premise is that agreement among judges can reveal the underlying quality signal by averaging out individual judge noise.

However, this premise breaks down when judges share the same nuisance sensitivity. LLM judges exhibit biases related to position, selection effects, and formatting artifacts \citep{zeng2024evaluating,li2024split,park2024offsetbias,ye2025justice,shi2025judging,li2025calibraeval}. When such biases are shared across judges, aggregation may reinforce rather than mitigate them. Judges can move together because an answer is genuinely better, but they can also move together because the answer is longer, better formatted, more fluent, or easier to parse. From observed scores alone, true quality and nuisance factors can be hard to distinguish.

Meanwhile, prior work has developed more sophisticated judge aggregation methods beyond majority voting, for example by estimating judge reliability, using higher-order agreement statistics, or modeling judge dependence \citep{ai2025beyond,balasubramanian2026dependence}. These tools improve aggregation by modeling how judge outputs interact with the latent quality signal. However, when shared nuisance factors induce score structure that resembles true quality, richer observational modeling can still amplify the wrong signal. More directly related to our work, \citet{zhao2026care} address shared nuisance factors by additionally introducing latent variables that capture their effects across judges. However, it still relies on observational judge outputs and structural assumptions such as sparsity and low rank. As a result, it does not provide a direct, controllable means of probing judges' sensitivity to specific nuisance attributes.

We introduce \voirdire{}, an \emph{intervention-based calibration} procedure for LLM-judge aggregation, named after \emph{voir dire}, the legal process of examining prospective jurors. \voirdire{} applies controlled interventions to the inputs shown to judges, using deterministic transformations or LLM-generated rewrites to perturb specified nuisance attributes while aiming to preserve the evaluation target. The same judges then score both the original and modified inputs. Paired score differences indicate which directions in judge-output space are sensitive to the targeted attributes. \voirdire{} attenuates these directions and selects aggregation weights that retain variation across items while penalizing sensitivity to the interventions.

Theoretically, we show that observational agreement alone can fail to identify the target signal, while paired original–rewrite responses enable estimation of intervention-sensitive nuisance variation, with error controlled by intervention quality and calibration sample size.

Empirically, we evaluate \voirdire{} on 12 datasets spanning continuous scoring, binary classification, and pairwise preference. Compared with the strongest existing judge-aggregation baseline on each dataset, \voirdire{} reduces MAE by up to \(26.8\%\) on continuous-score tasks and improves accuracy by up to \(5.3\) percentage points on binary/preference tasks. Controlled experiments further show that \voirdire{} reduces aggregate sensitivity to targeted nuisance changes. We also examine the effect of calibration-pair budget and interpret how calibration changes judge scores and aggregation weights. In summary, our contributions include:

\begin{enumerate}[leftmargin=*,itemsep=0.15em,topsep=0.25em]
    \item We introduce \voirdire{}, an intervention-based calibration method that uses controlled perturbations to mitigate shared nuisance sensitivity in LLM-judge aggregation.
    \item We characterize the limitations of observational judge agreement and establish conditions under which paired interventions enable accurate estimation of nuisance-sensitive directions.
    \item Across 12 datasets spanning continuous scoring, binary classification, and pairwise preference, \voirdire{} reduces MAE by up to \(26.8\%\) and improves accuracy by up to \(5.3\) percentage points over existing aggregation methods.
\end{enumerate}

\section{Related Work}\label{sec:related-work}

\paragraph{LLM-as-a-judge evaluation.}
LLM judges are widely used to evaluate open-ended generation because they scale beyond manual annotation and can express task-specific rubrics \citep{liu2023geval,zheng2023judging,wang2024pandalm,zhu2025judgelm}. However, their judgments can depend on presentation, verbosity, position, selection effects, and other nuisance factors rather than only the intended evaluation target \citep{zeng2024evaluating,li2024split,park2024offsetbias,ye2025justice,shi2025judging,li2025calibraeval}. Aggregating multiple judges can reduce individual noise, but it does not necessarily eliminate shared bias.

\begin{figure*}[!t]
    \centering
    \includegraphics[width=\textwidth]{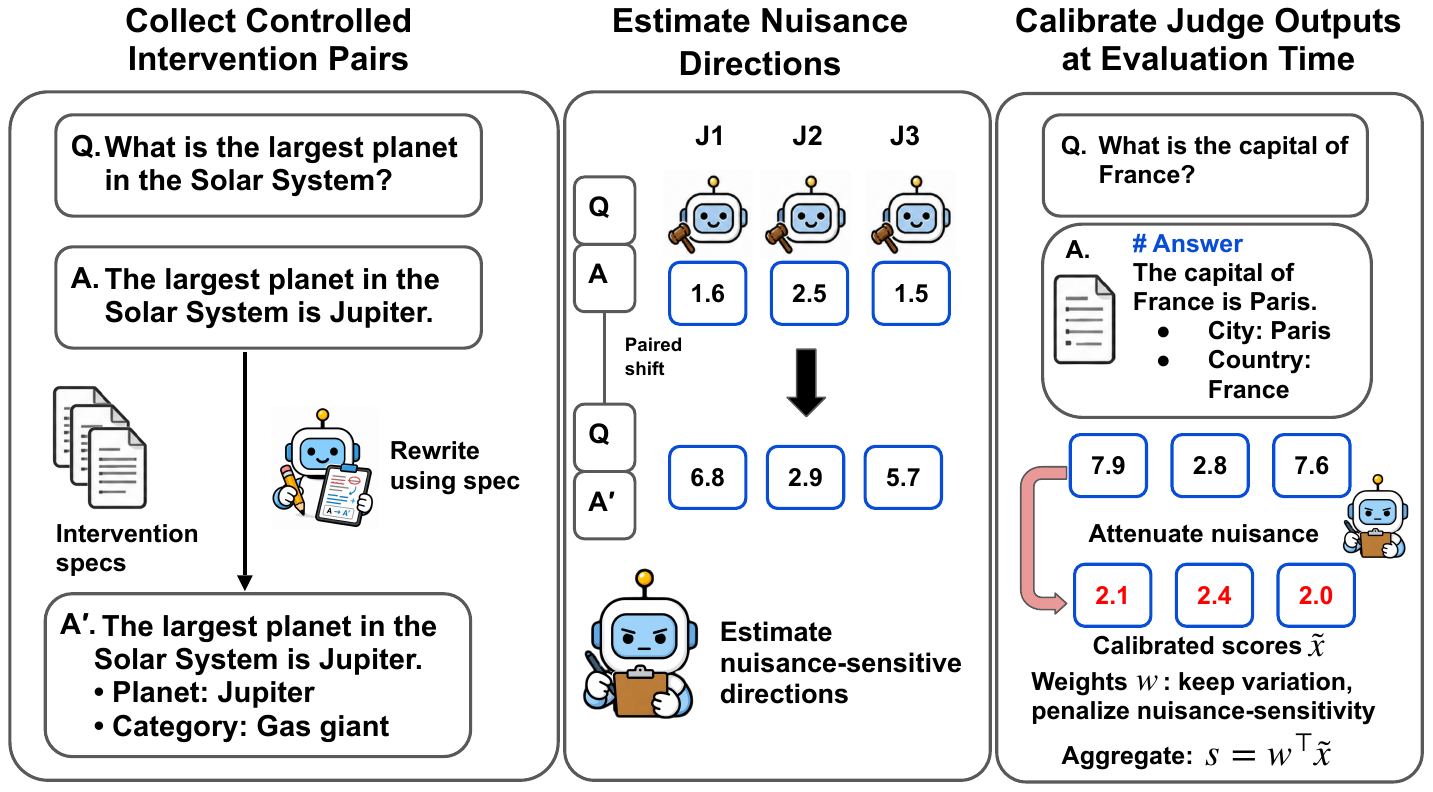}
    \vspace{-1.5em}
    \caption{\voirdire{} framework. Controlled interventions give paired judge-score shifts, pooled into \(\bar G\). At evaluation time, judge outputs are attenuated along its directions (\cref{eq:spectral-calibration}) and weighted to retain item variation while penalizing sensitivity to the shifts (\cref{eq:aggregation-objective}).}
    \label{fig:voirdire-framework}
\end{figure*}

\paragraph{Judge aggregation.}
Classical crowdsourcing and weak-supervision methods infer latent labels from multiple noisy sources by modeling source reliability, dependence, or latent structure \citep{dawid1979maximum,karger2011iterative,ratner2017snorkel,shin2022universalizing}. These approaches naturally extend to aggregating LLM judges. Recent LLM-jury methods use richer observational statistics, including higher-order agreement signals such as Inverse Surprising Popularity (ISP) and dependence-aware Ising models \citep{ai2025beyond,balasubramanian2026dependence}. However, these methods do not explicitly account for nuisance factors shared across judges. More directly related to our work, CARE \citep{zhao2026care} models such shared factors using latent variables inferred from observed judge outputs. Still, its identification relies on observational outputs and structural assumptions such as sparsity and low rank. \voirdire{} instead uses controlled interventions to directly measure which judge-output directions respond to nuisance changes.

\paragraph{Counterfactual interventions and causal analysis.}
Causal inference distinguishes observational association from the effects revealed by controlled interventions \citep{pearl2009causality}. In NLP, this view has been used to estimate the causal effects of textual concepts on model behavior \citep{feder2021causalm,abraham2022cebab,feder2022causal} and to make models invariant to spurious attributes using counterfactual pairs \citep{kaushik2020learning,veitch2021counterfactual}; foundation models can now generate such counterfactuals at scale \citep{wu2021polyjuice,chen2023disco,feder2023data}. \voirdire{} brings this causal view to judge aggregation.
\section{\voirdire{}: Calibrating Judge Aggregation with Interventions}
\label{sec:method}

\voirdire{} calibrates LLM-judge aggregation using controlled interventions. It first constructs original--rewrite pairs designed to preserve the evaluation target while varying specified nuisance attributes, such as formatting, style, or verbosity. The same judges evaluate both inputs, and their paired score differences reveal nuisance-sensitive directions in judge-output space. \voirdire{} uses these directions to calibrate judge scores and determine intervention-aware aggregation weights. \Cref{fig:voirdire-framework} gives an overview, and \cref{alg:voirdire} summarizes the procedure.

\paragraph{Setup.}
We consider a panel of \(p\) LLM judges. For item \(i\), the judges see a judge input \(u_i\) and produce a judge-output vector \(x_i=(x_{i1},\ldots,x_{ip})\). Entries may be scalar scores, binary labels, or preference votes. A standard aggregator, such as majority voting, uses only the unperturbed judge-output matrix \(X=\{x_i\}_{i=1}^n\) to predict a target score or label.

\paragraph{Constructing intervention pairs.}
We begin with a pool of candidate intervention families targeting nuisance attributes such as formatting, verbosity, and candidate position. Each family \(a\) specifies three components: the nuisance attribute and input span to modify, a modification rule, and preservation constraints. The modification rule is either deterministic (e.g., adding markup or swapping two candidates verbatim) or an instruction to an LLM rewrite model. For example, a concision family instructs the model to remove redundancy while preserving the original claims, sentiment, numbers, and other content constraints.

In our experiments, we use LLM assistants to propose candidate intervention families based on dataset examples. We apply each intervention to a subset of inputs and retain only pairs that pass programmatic preservation checks. The same judges score each retained modified input, yielding \(x_i^{(a)}\) for \(i\in\mathcal P_a\). Pair generation and preservation checking use no ground-truth labels. Full construction and selection details appear in \cref{app:experiment-details}.

\paragraph{Estimating nuisance-sensitive directions.}
Given \(m\) calibration pairs from each of the \(k\) selected intervention families, we compute the paired score differences \(\Delta_i^{(a)}=x_i^{(a)}-x_i\) for \(i\in\mathcal P_a\). These vectors record how the judges respond to each intervention. When the evaluation target remains unchanged, they capture nuisance-sensitive variation together with judge noise. We pool the differences into a second-moment matrix\footnote{Note that we do not subtract the mean shift: a consistent increase or decrease in scores can itself reflect bias, such as a position preference, and centering might remove this signal (\cref{prop:paired-difference}).}:
\begingroup\setlength{\abovedisplayskip}{3pt}\setlength{\belowdisplayskip}{3pt}\setlength{\abovedisplayshortskip}{1pt}\setlength{\belowdisplayshortskip}{1pt}
\begin{equation}\label{eq:family-shift-covariance}
 \bar G=\frac{1}{km}\sum_{a=1}^{k}\sum_{i\in\mathcal P_a}
 \Delta_i^{(a)}\Delta_i^{(a)\top},
 \qquad |\mathcal P_a|=m,\quad M=km.
\end{equation}\endgroup
Each pair contributes equally to \(\bar G\in\mathbb R^{p\times p}\). Its entry \(\bar G_{j\ell}\) captures how judges \(j\) and \(\ell\) respond together, and its leading eigenvectors identify the directions along which the interventions produce the largest score changes. These directions capture nuisance sensitivity, although judge noise and imperfections in the interventions can affect their estimation.

\paragraph{Attenuating sensitive directions.}
Using the estimated directions, we calibrate each judge-output vector by attenuating components along directions that respond more strongly to the interventions:
\begingroup\setlength{\abovedisplayskip}{3pt}\setlength{\belowdisplayskip}{3pt}\setlength{\abovedisplayshortskip}{1pt}\setlength{\belowdisplayshortskip}{1pt}
\begin{equation}\label{eq:spectral-calibration}
 \widetilde x_i=x_i-\alpha\frac{\bar Gx_i}{\lambda_{\max}(\bar G)},
 \qquad \alpha\in[0,1].
\end{equation}\endgroup
Here \(\lambda_{\max}(\bar G)\) is the largest eigenvalue, and \(\alpha\) is a hyperparameter controlling attenuation strength. Normalizing by \(\lambda_{\max}(\bar G)\) ensures that the most sensitive direction is attenuated by a fraction \(\alpha\), while less sensitive directions are attenuated less.

\paragraph{Intervention-aware aggregation.}
After calibrating the judge outputs, we aggregate them using weights that retain variation across items while penalizing sensitivity to the interventions:
\begin{equation}\label{eq:aggregation-objective}
w_\star\in\arg\max_{w\ne0}
\frac{\widehat{\operatorname{Var}}_i(w^\top\widetilde x_i)}
{w^\top\bar Gw+\rho\|w\|_2^2}.
\end{equation}
The numerator rewards variation across the calibrated items, while \(w^\top\bar Gw\) penalizes weights that are more sensitive to nuisance changes. The hyperparameter \(\rho>0\) controls the tradeoff between retaining variation and reducing intervention sensitivity. We solve the resulting generalized eigenvalue problem, then orient and normalize the solution to obtain the final aggregation weights \(w\). The aggregate score is \(s_i=w^\top\widetilde x_i\). Implementation details appear in \cref{app:current-protocol}.

\begin{algorithm}[t]
    \caption{\voirdire{}: calibration and intervention-aware aggregation}
    \label{alg:voirdire}
    \begin{algorithmic}[1]
    \REQUIRE Judge outputs \(x_1,\ldots,x_n\in\mathbb R^p\);
    paired outputs \(\{(x_i,x_i^{(a)})\}_{i\in\mathcal P_a}\)
    for \(k\) selected intervention families, with
    \(|\mathcal P_a|=m\) and \(M=km\);
    validation-selected hyperparameters \((\alpha,\gamma)\).
    \ENSURE Aggregate scores \(s_1,\ldots,s_n\).

    \STATE Compute paired score differences for each family \(a\)
    and item \(i\in\mathcal P_a\):\\
    \hspace*{1.2em}
    \(\Delta_i^{(a)}\leftarrow x_i^{(a)}-x_i\).

    \STATE Estimate the pooled second-moment matrix:\\
    \hspace*{1.2em}
    \(\bar G\leftarrow\frac{1}{M}\sum_{a=1}^{k}
    \sum_{i\in\mathcal P_a}
    \Delta_i^{(a)}\Delta_i^{(a)\top}\).
    \hfill \cref{eq:family-shift-covariance}

    \STATE Attenuate nuisance-sensitive components for all items:\\
    \hspace*{1.2em}
    \(\widetilde x_i\leftarrow
    x_i-\alpha\,\bar Gx_i/\lambda_{\max}(\bar G)\),
    \(i=1,\ldots,n\).
    \hfill \cref{eq:spectral-calibration}

    \STATE Set \(\rho\leftarrow\gamma\operatorname{tr}(\bar G)/p\)
    and solve for intervention-aware weights:\\
    \hspace*{1.2em}
    \(w_\star\in\arg\max_{w\ne0}
    \frac{\widehat{\operatorname{Var}}_i(w^\top\widetilde x_i)}
    {w^\top\bar Gw+\rho\|w\|_2^2}\).
    \hfill \cref{eq:aggregation-objective}

    \STATE Orient and normalize \(w_\star\) to obtain \(w\)
    (\cref{app:current-protocol}).

    \STATE \textbf{return}
    \(s_i=w^\top\widetilde x_i\), \(i=1,\ldots,n\).
    \end{algorithmic}
    \end{algorithm}
\section{Theoretical Analysis}\label{sec:theory}
We theoretically analyze \voirdire{}, showing that observational judge agreement alone may not distinguish target quality from shared nuisance factors, whereas paired interventions reveal nuisance-sensitive variation and enable estimation of the calibration matrix.

\paragraph{Setup.}
We model judge outputs with target-relevant factors \(q_i\), nuisance factors \(c_i\), and noise:
\begin{equation}\label{eq:minimal-intervention-model}
\begin{aligned}
x_i^{(0)} &= U_Qq_i+U_Cc_i+\varepsilon_i^{(0)},\\
x_i^{(a)} &= U_Qq_i+U_C(c_i+\delta_i^{(a)})
             +\varepsilon_i^{(a)},
\end{aligned}
\end{equation}
where \(x_i^{(0)}=x_i\in\mathbb R^p\) denotes the original judge-output vector, and \(x_i^{(a)}\) denotes the outputs of the same judges after an intervention from family \(a\). The matrices \(U_Q\) and \(U_C\) map target-relevant and nuisance factors to judge scores, respectively. We first consider an ideal setting in which interventions change nuisance factors by \(\delta_i^{(a)}\) while leaving \(q_i\) unchanged.

\paragraph{Why agreement alone is insufficient.}
Shared variation in judge scores need not reveal whether judges respond to target quality or nuisance attributes; Theorem~\ref{thm:observational-nonidentifiability} makes this explicit.

\begin{theorem}[Observational non-identifiability; simplified]
    \label{thm:observational-nonidentifiability}
    In the noiseless model \(x=uq+vc\), with orthonormal \(u,v\in\mathbb R^p\) and \((q,c)\sim N(0,I_2)\), the target direction \(u\) is not identifiable from the distribution of \(x\): rotating \(u\) and \(v\) within their span leaves this distribution unchanged.
\end{theorem}

Thus, even with unlimited observations, an aggregator cannot determine which direction represents quality in this model without additional information or assumptions.

\paragraph{What paired interventions reveal.}
Paired interventions address this ambiguity by holding target quality fixed while changing nuisance attributes. Under \cref{eq:minimal-intervention-model}, the target component cancels: \( \Delta_i^{(a)}=x_i^{(a)}-x_i^{(0)} =U_C\delta_i^{(a)}+\xi_i^{(a)}, \) where \(\xi_i^{(a)}\) is the paired judge noise. If this noise is mean-zero and independent of \(\delta_i^{(a)}\), with finite second moments, the pooled population second moment is \(G=L_C+\Omega\), where \(L_C=\frac1k\sum_a \mathbb E[(U_C\delta_i^{(a)})(U_C\delta_i^{(a)})^\top]\) captures intervention-induced nuisance variation and \(\Omega\) is the pooled noise covariance.
\begin{proposition}[Calibration-matrix estimation; simplified]
\label{prop:paired-difference}
Suppose \(\|G-L_C\|_{\mathrm{op}}\le\rho_{\mathrm{err}}\) and \(\|\bar G-G\|_{\mathrm{op}}\le\epsilon_M\). For nonzero \(L_C,\bar G\), let \(T=L_C/\lambda_{\max}(L_C)\) and \(\widehat T=\bar G/\lambda_{\max}(\bar G)\). Then
\[
\|\widehat T-T\|_{\mathrm{op}}
\le
\frac{2(\rho_{\mathrm{err}}+\epsilon_M)}
{\lambda_{\max}(L_C)}.
\]
\end{proposition}
\Cref{prop:paired-difference} shows that accurate calibration requires estimation and systematic errors to be small relative to the nuisance signal. We further analyze how calibration sample size affects estimation error (\cref{thm:voirdire-sample}) and how imperfect interventions affect nuisance recovery. Full proofs and additional theoretical results appear in \cref{app:theory-proofs}, with supporting synthetic experiments in \cref{app:synthetic-theory-stress}.

\begin{table*}[t]
\centering
\small
\setlength{\tabcolsep}{3pt}
\begin{tabular}{lrrrrrr}
\toprule
Dataset & MV & AVG & ISP & PCA & CARE-SVD & \voirdire{} \\
\midrule
Yelp & $1.016$ & $1.089$ & $1.028$ & $1.054$ & $0.694$ & $\mathbf{0.508}$ \\
SummEval & $0.819$ & $0.753$ & $0.832$ & $0.738$ & $0.729$ & $\mathbf{0.701}$ \\
Summarize from Feedback & $1.434$ & $1.453$ & $1.585$ & $1.764$ & $1.392$ & $\mathbf{1.346}$ \\
HelpSteer2 & $0.998$ & $0.981$ & $1.008$ & $0.992$ & $0.998$ & $\mathbf{0.951}$ \\
ASSET & $31.238$ & $33.701$ & $31.885$ & $28.888$ & $27.629$ & $\mathbf{26.930}$ \\
Review-5K & $2.581$ & $2.483$ & $2.585$ & $2.101$ & $1.957$ & $\mathbf{1.862}$ \\
\bottomrule
\end{tabular}
\caption{Continuous-score aggregation: mean MAE ($\downarrow$).}
\label{tab:continuous-real}
\end{table*}

\begin{table*}[t]
\centering
\small
\setlength{\tabcolsep}{3pt}
\begin{tabular}{lrrrrrrr}
\toprule
Dataset & MV & AVG & ISP & PCA & CARE-SVD & CARE-Tensor & \voirdire{} \\
\midrule
Chatbot Arena & $0.517$ & $0.551$ & $0.571$ & $0.544$ & $0.572$ & $0.596$ & $\mathbf{0.646}$ \\
RewardBench 2 & $0.577$ & $0.596$ & $0.569$ & $0.595$ & $0.578$ & $0.605$ & $\mathbf{0.658}$ \\
LIAR2 & $0.678$ & $0.665$ & $0.678$ & $0.661$ & $0.670$ & $0.668$ & $\mathbf{0.686}$ \\
HelpSteer3 & $0.490$ & $0.517$ & $0.556$ & $0.520$ & $0.592$ & $0.572$ & $\mathbf{0.610}$ \\
CivilComments & $0.702$ & $0.715$ & $0.716$ & $0.714$ & $\mathbf{0.747}$ & $0.731$ & $0.739$ \\
MT-Bench (human) & $0.600$ & $0.613$ & $0.600$ & $0.599$ & $0.594$ & $0.581$ & $\mathbf{0.624}$ \\
\bottomrule
\end{tabular}
\caption{Binary/preference aggregation: mean accuracy ($\uparrow$).}
\label{tab:binary-real}
\end{table*}

\section{Experiments}\label{sec:experiments}
We evaluate \voirdire{} across continuous-score, binary, and preference LLM-judge aggregation setups. The experiments address the following topics:
\begin{itemize}[nosep,topsep=0pt,leftmargin=*]
    \item \textbf{Improving judge aggregation (\cref{subsec:exp_main_agg}):} We compare \voirdire{} with existing aggregation methods across the three task types.
    \item \textbf{Ablating score calibration and aggregation (\cref{subsec:gain-decomposition}):} We examine how score calibration and intervention-aware
aggregation individually contribute to performance.
    \item \textbf{Judge-level interpretation (\cref{subsec:exp_interpretation}):} We examine how intervention sensitivity relates to judge reliability and aggregation weights.
    \item \textbf{Calibration efficiency (\cref{subsec:pairs-ablation}):} We measure how performance changes with the number of intervention pairs.
    \item \textbf{Targeted nuisance removal (\cref{subsec:injected-bias}):} We test whether \voirdire{} reduces score shifts under rewrites that trigger known judge biases.
    \item \textbf{Transfer across related nuisances (\cref{subsec:intervention-transfer}):} We test whether calibration transfers to related nuisance variations not used during fitting.
\end{itemize}

\paragraph{Datasets and metrics.}
We evaluate on six continuous-score and six binary/preference datasets. We report mean absolute error (MAE; lower is better) for continuous scoring and accuracy (higher is better) for binary classification and pairwise preference tasks. Dataset details are in \cref{app:datasets}.

\paragraph{LLM judges.}
Continuous-score panels contain 11 open-weight LLM judges and binary/preference panels 20, with model sizes ranging from 0.6B to 14B parameters. Each judge reports on the dataset-native score scale. Following the judge-output convention in CARE \citep{zhao2026care}, judges first produce scalar scores for binary classification and pairwise preference tasks. These scores are then converted to binary labels or preferences using fixed thresholds and evaluated by accuracy. Details are in \cref{appsec:judge-models,appsec:judge-prompts}.

\paragraph{Interventions.}
We use LLM assistants to help identify task-specific nuisance attributes and design interventions, following \citet{adila2024zero}. The interventions vary formatting, verbosity, or candidate presentation while aiming to preserve the evaluation target. We generate modified inputs through deterministic transformations or rewrites by one of the judge models, \texttt{Llama-3.1-8B-Instruct}. Pair generation requires no ground-truth labels; family selection and tuning of \((\alpha,\gamma)\) use the validation set. Full specifications, prompts, and selection details appear in \cref{app:experiment-details}.

\subsection{Improving Aggregation with Interventions}
\label{subsec:exp_main_agg}

\paragraph{Setup.}
We compare \voirdire{} with majority voting (MV), score averaging (AVG), PCA aggregation, CARE \citep{zhao2026care}, and Inverse Surprising Popularity (ISP) \citep{ai2025beyond}. We use CARE-SVD for continuous scoring and both CARE-SVD and CARE-Tensor for binary classification and pairwise preference. All methods use the same judges and evaluation items across five random splits per dataset. For \voirdire{}, we use \(m=32\) intervention pairs per family. Protocol details and additional results in \cref{app:experiment-details,app:full-experiment-results}, respectively.

\paragraph{Results.}
Tables~\ref{tab:continuous-real} and~\ref{tab:binary-real} show that \voirdire{} improves on existing aggregation methods across task types. The largest MAE reduction is on Yelp, at \(26.8\%\) relative to the strongest baseline, and the largest accuracy gains are on RewardBench~2 (\(5.3\) pp) and Chatbot Arena (\(5.1\) pp). These improvements suggest that intervention-based calibration can enhance judge aggregation compared with existing observational methods.

\subsection{Ablating Score Calibration and Aggregation}
\label{subsec:gain-decomposition}
\voirdire{} uses interventions for score calibration and intervention-aware aggregation. We examine the contribution of each step.

\paragraph{Setup.} Following the setup of the previous subsection, we compare PCA and intervention-aware aggregation on raw and calibrated scores. The PCA baseline corresponds to \(\alpha=0\) and \(\rho\to\infty\) in \cref{eq:aggregation-objective}. For intervention-aware aggregation on raw scores, we set \(\alpha=0\) and re-tune \(\rho\) on validation. Combining score calibration and intervention-aware aggregation gives \voirdire{}.

\begin{figure}[!tbp]
\centering
\includegraphics[width=\linewidth]{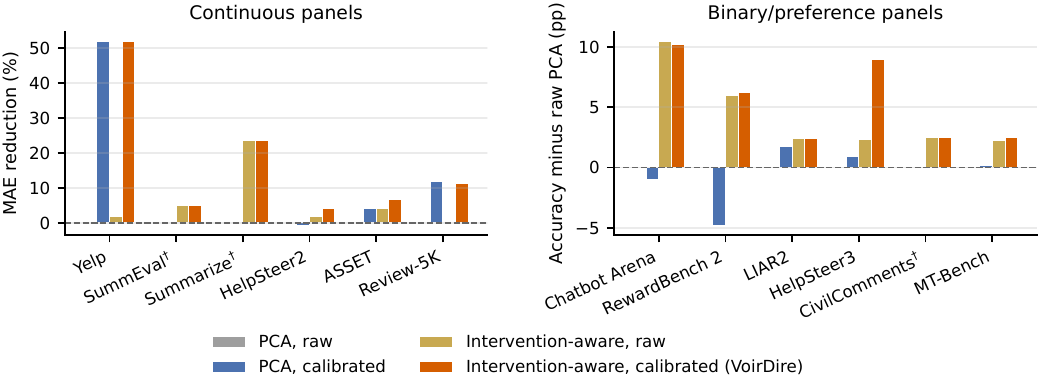}
\caption{Contributions of calibration and aggregation relative to raw PCA: relative MAE reduction (\%) on continuous tasks and accuracy gain (percentage points) on binary/preference tasks. Higher is better. \(\dagger\): \(\alpha=0\), so calibration leaves scores unchanged.}
\label{fig:gain-factorial}
\end{figure}

\paragraph{Results.} \Cref{fig:gain-factorial} shows that both steps contribute, with their relative benefits varying across datasets. On continuous-score tasks, score calibration provides substantial gains on datasets such as Yelp and ASSET. On binary/preference tasks, intervention-aware aggregation alone improves over PCA on all six datasets, while combining both steps provides further gains on datasets such as HelpSteer3. Additional ablations on other aggregators appear in \cref{app:gain-decomposition}.

\subsection{How Interventions Diagnose and Reweight Judges}
\label{subsec:exp_interpretation}
Beyond improving aggregation, controlled interventions can help diagnose individual judges by revealing their sensitivity to nuisance attributes. We examine whether this sensitivity relates to judge reliability and how it influences the aggregation weights learned by \voirdire{}.

\paragraph{Setup.}
For each judge in the panels of \cref{subsec:exp_main_agg}, we measure \emph{rewrite sensitivity} as the root mean square score shift across calibration pairs, \(\sqrt{\bar G_{jj}}\), normalized by the standard deviation of its original scores to account for differences in score scale. We measure \emph{reliability} as the Spearman correlation between the judge's original scores and ground truth, capturing how well its judgments align with the evaluation target. Comparing these metrics tests whether nuisance sensitivity, measured without ground-truth labels, is associated with judge reliability. We also examine the aggregation weights \(w_j\) from \cref{eq:aggregation-objective} to assess how \voirdire{} uses this sensitivity information.

\begin{figure}[!tbp]
    \centering
    \includegraphics[width=\linewidth]{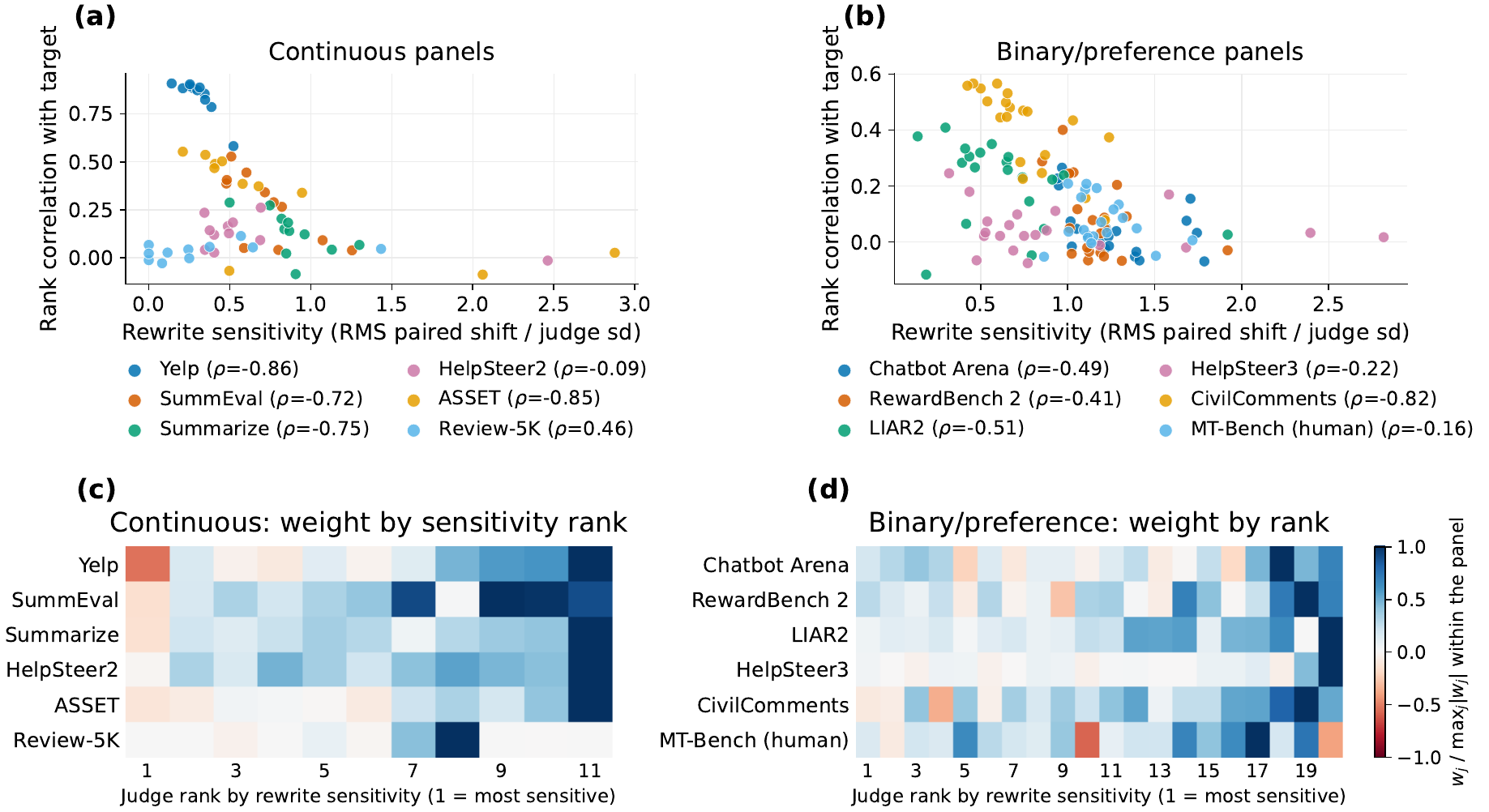}
    \caption{Rewrite sensitivity, reliability, and aggregation weights. (a, b) One point per judge and panel; legends report the within-panel Spearman correlation between sensitivity and reliability. (c, d) Weights divided by each panel's largest absolute weight, with judges ranked by sensitivity (1: most sensitive; blue: positive, red: negative). Empty ranks indicate fewer judges.}
    \label{fig:judge-calibration-main}
\end{figure}

\paragraph{Results.}
\Cref{fig:judge-calibration-main} (a), (b) show that judges with greater rewrite sensitivity tend to be less reliable, although the relationship varies across datasets. Panels (c) and (d) show that \voirdire{} generally assigns lower weights to more sensitive judges.

\subsection{Ablation on the Number of Intervention Pairs}
\label{subsec:pairs-ablation}
\begin{wrapfigure}{r}{0.38\textwidth}
    \vspace{-6pt}
    \centering
    \captionsetup{font=small,skip=2pt}
    \includegraphics[width=\linewidth]{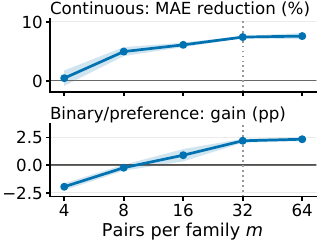}
    \caption{Mean gain over the strongest baseline. Bands: one SE over five splits; dotted line: \(m{=}32\).}
    \label{fig:pairs-ablation}
    \vspace{-10pt}
\end{wrapfigure}
We examine how the number of intervention pairs affects the performance of \voirdire{}.

\paragraph{Setup.}
We vary the number of calibration pairs per family \(m\in\{4,8,16,32,64\}\), keeping the intervention families of \cref{subsec:exp_main_agg} and re-selecting \((\alpha,\rho)\) on validation at every \(m\). We report improvements over each dataset's strongest baseline, averaged separately across the six continuous-score datasets (relative MAE reduction) and the six binary/preference datasets.

\paragraph{Results.}
\Cref{fig:pairs-ablation} shows that \voirdire{} improves aggregation with a small calibration budget. Eight pairs per family recover most of the gain on continuous-score tasks, and gains on both task types level off between 32 and 64 pairs. With 32 pairs per family, \voirdire{} achieves an average MAE reduction of \(7.4\%\) and an accuracy gain of \(2.2\) percentage points over each dataset's strongest baseline. Per-dataset results appear in \cref{app:pairs-per-family-budget}.

\subsection{Targeted Nuisance Removal}
\label{subsec:injected-bias}
We further evaluate whether \voirdire{} reduces sensitivity to known nuisance changes in a controlled setting, directly testing whether it achieves its intended effect.



\begin{table}[t]
\centering
\small
\caption{Examples of targeted nuisance rewrites (left)
and mean absolute score shifts (right, $\downarrow$).}
\label{tab:injected-bias-m32}
\vspace{3pt}

\begin{minipage}[t]{0.54\linewidth}
\vspace{0pt}
\footnotesize
\setlength{\tabcolsep}{3pt}
\renewcommand{\arraystretch}{1.15}
\begin{tabularx}{\linewidth}{@{}lX@{}}
\toprule
\textbf{Type} & \textbf{Example} \\
\midrule
Original &
Night and day occur due to the Earth's rotation on its axis. \\
\addlinespace[3pt]
Beauty &
Night and day occur due to the
\textbf{Earth's rotation} on its axis.
When facing the Sun~\sun{}, it experiences
\textbf{daylight}. \ldots \\
\addlinespace[3pt]
Authority &
\ldots{} This rotation happens over a period of
24 hours, creating our cycle of day and night
(National Geographic, 2019). \\
\bottomrule
\end{tabularx}
\end{minipage}
\hfill
\begin{minipage}[t]{0.43\linewidth}
\vspace{0pt}
\centering
\small
\setlength{\tabcolsep}{3pt}
\begin{tabular}{@{}lrr@{}}
\toprule
Method & Beauty & Authority \\
\midrule
MV & $0.919$ & $0.824$ \\
AVG & $0.506$ & $0.325$ \\
ISP & $0.891$ & $0.733$ \\
PCA & $1.283$ & $0.876$ \\
CARE-SVD & $0.418$ & $0.286$ \\
\voirdire{} & $\mathbf{0.304}$ & $\mathbf{0.234}$ \\
\bottomrule
\end{tabular}

\end{minipage}

\vspace{-5pt}
\end{table}

\paragraph{Setup.}
We use the judgment-bias benchmark of \citet{chen-etal-2024-humans}, which provides original--modified answer pairs with GPT-4 rewrites targeting rich formatting (beauty) or added references (authority), along with 1--10 judge scores. These rewrites can also rephrase or expand the original answers. We measure sensitivity as the mean absolute difference
between aggregate scores on original and rewritten answers. Lower values indicate greater robustness to the targeted nuisance changes.


\paragraph{Results.}
\voirdire{} achieves the lowest sensitivity to both biases (\cref{tab:injected-bias-m32}), reducing mean absolute score shifts by approximately \(27.3\%\) for beauty and \(18.2\%\) for authority relative to CARE-SVD, the strongest baseline. Additional experimental details and results appear in \cref{app:injected-bias,app:injected-bias-results}.

\subsection{Transfer Across Related Nuisances}
\label{subsec:intervention-transfer}
Identifying every nuisance variation can be costly. However, related rewrites may affect judge scores in similar ways, so we expect \voirdire{} to generalize beyond the specific interventions used for calibration. We test this in a controlled setting.
\begin{wraptable}{R}{0.48\textwidth}
    \vspace{-6pt}
    \captionsetup{position=top,skip=3pt}
\centering\small
\setlength{\tabcolsep}{3pt}
\caption{Percentage of target-family gain retained ($\uparrow$).}
\label{tab:intervention-transfer}
\begin{tabular}{@{}llrr@{}}
\toprule
\multicolumn{2}{c}{Calibration} & \multicolumn{2}{c}{Evaluation} \\
\cmidrule(lr){1-2}\cmidrule(l){3-4}
Concept & Variant & Beauty & Authority \\
\midrule
Beauty & Numbered & $77.1\%$ & $91.3\%$ \\
 & Panels & $81.1\%$ & $102.5\%$ \\
 & Icons & $72.0\%$ & $100.3\%$ \\
\midrule
Authority & Expert & $60.4\%$ & $100.5\%$ \\
 & Source header & $74.6\%$ & $99.2\%$ \\
 & Footnote & $63.9\%$ & $98.0\%$ \\
\bottomrule
\end{tabular}

    \vspace{-6pt}
\end{wraptable}

\paragraph{Setup.}
Following \cref{subsec:injected-bias}, we calibrate on three beauty variants (numbered structure, minimalist panels, and topic-specific icons) and three authority variants (expert attribution, a source header, and a fabricated footnote), then evaluate on the original beauty and authority rewrites. We report sensitivity reduction relative to PCA,
normalized so that calibration on the original beauty or authority rewrites achieves \(100\%\). Details appear in \cref{app:chen-transfer-details}.

\paragraph{Results.}
\Cref{tab:intervention-transfer} shows that calibration retains much of its benefit on unseen rewrites. Notably, beauty variants transfer strongly even to authority, while authority variants also provide some benefit on beauty. This suggests that distinct nuisance attributes may affect overlapping directions in judge-output space, enabling transfer across them.
\section{Conclusion}
We introduced \voirdire{}, which uses controlled interventions to mitigate shared nuisance sensitivity in LLM-judge aggregation. Across 12 datasets, \voirdire{} improves aggregation performance and reduces sensitivity to targeted nuisance changes. Calibration also transfers to related, unseen variations, suggesting that exact rewrite matches are unnecessary. These findings highlight the potential of counterfactual interventions for more reliable LLM-based evaluation.

\paragraph{Limitations and future work.}
\voirdire{} depends on the quality of its interventions: they should capture relevant nuisance variation while preserving the evaluation target, which is difficult to guarantee. Discovering effective interventions also adds cost, and calibration requires additional inference to generate rewrites and score modified inputs, although deterministic transformations avoid the rewrite-generation cost. More broadly, paired interventions could extend beyond aggregation: agents could actively search for interventions that expose previously unknown judge biases, and the same paired signals could train judges and reward models to be invariant to surface features such as length and formatting, reducing reward hacking in downstream optimization.

\clearpage
\subsection*{Ethics statement}
The main risk is over-trust in calibrated automatic evaluation. \voirdire{} attenuates nuisance sensitivities exposed by the chosen interventions, but it does not guarantee unbiased judgments or remove all evaluation artifacts. For private or high-stakes evaluations, rewrite generation and judge calls should be run with appropriate data-protection safeguards, since the method sends original and rewritten inputs to LLMs.

\subsection*{AI use statement}
We used AI coding and writing assistants (Claude Code, Codex, Fable, and ChatGPT Pro) throughout this project. They assisted with implementing and running experimental, evaluation, and plotting code; proposing nuisance attributes and intervention designs; developing and checking the theoretical analysis; and drafting and editing the manuscript. The authors directed all AI-assisted work, reviewed its outputs, and take full responsibility for the content.

\bibliography{custom}
\bibliographystyle{iclr2027_conference}

\appendix
\crefalias{section}{appendix}
\crefalias{subsection}{appendix}
\crefalias{subsubsection}{appendix}
\section*{Appendix Overview}

This appendix starts with a glossary of notation in \cref{app:glossary}. It then provides experimental details in \cref{app:experiment-details}, including datasets, experiment protocols, compute resources, judge prompts, and intervention specifications. Next, \cref{appsec:theory-proofs} contains proofs and additional theoretical results. We then present synthetic experiments that study calibration behavior in controlled settings (\cref{app:synthetic-theory-stress}). Finally, \cref{app:full-experiment-results} reports additional experimental results.

\clearpage
\section{Glossary}\label{app:glossary}

The notation used throughout the paper is summarized in \cref{tab:glossary}.

\vspace{-0.5em}
\begin{center}
\setlength{\tabcolsep}{5pt}
\renewcommand{\arraystretch}{0.95}
\captionof{table}{Glossary of variables, symbols, and abbreviations.}
\label{tab:glossary}

\begin{tabular}{p{0.24\linewidth}p{0.68\linewidth}}
\toprule
Symbol & Definition \\
\midrule
\(i,j,a\) & Item, judge, and intervention-family indices. \\
\(n,p,k\) & Numbers of original items, judges, and selected families. \\
\(m,m_a,M\) & Pairs per family and total budget:
\(M=km\), or \(M=\sum_a m_a\) for unequal budgets. \\
\(\alpha,\gamma,\rho\) & Attenuation strength, regularization multiplier,
and ridge parameter \(\rho=\gamma\operatorname{tr}(\bar G)/p\). \\
\(u_i\) & Judge input for item \(i\). \\
\(x_i,x_i^{(0)}\) & Original judge-output vector;
\(x_i^{(0)}=x_i\in\mathbb R^p\). \\
\(X\) & Matrix of original judge outputs. \\
\(x_i^{(a)}\) & Judge-output vector after intervention \(a\). \\
\(\mathcal P_a\) & Calibration item indices for family \(a\). \\
\(\Delta_i^{(a)},z_i^{(a)}\) & Paired score difference
\(x_i^{(a)}-x_i^{(0)}\). \\
\(\bar G\) & Empirical second moment of pooled paired differences
(\cref{eq:family-shift-covariance}). \\
\(G,L_C\) & Population second moments of paired differences
and intervention-induced nuisance variation, respectively. \\
\(T,\widehat T\) & Ideal and empirical normalized operators:
\(T=L_C/\lambda_{\max}(L_C)\),
\(\widehat T=\bar G/\lambda_{\max}(\bar G)\). \\
\(\eta\) & Operator-estimation error bound:
\(\|\widehat T-T\|_{\mathrm{op}}\le\eta\). \\
\(\widetilde x_i\) & Calibrated output
\((I-\alpha\widehat T)x_i\). \\
\(\widehat\Sigma_\alpha\) & Empirical covariance of calibrated outputs. \\
\(w\) & Oriented and normalized aggregation weights
(\cref{eq:aggregation-objective}). \\
\(s_i\) & Aggregate score \(w^\top\widetilde x_i\). \\
\(q_i,c_i,\delta_i^{(a)}\) & Target factors, nuisance factors,
and intervention-induced changes in nuisance factors. \\
\(U_Q,U_C\) & Loading matrices for target and nuisance factors. \\
\(d_C\) & Nuisance-subspace dimension. \\
\(\widehat U_C,P_{U_C}\) & Leading \(d_C\) eigenvectors of \(\bar G\)
and orthogonal projector onto \(\operatorname{range}(U_C)\). \\
\(\varepsilon_i^{(a)},\xi_i^{(a)}\) & Judge noise and paired noise
\(\xi_i^{(a)}=\varepsilon_i^{(a)}-\varepsilon_i^{(0)}\). \\
\(M_a\) & Family-specific second moment
\(\mathbb E[z_i^{(a)}z_i^{(a)\top}]\). \\
\(A_a,\Omega_a\) & Nuisance-shift second moment
\(\mathbb E[\delta_i^{(a)}\delta_i^{(a)\top}]\)
and paired-noise covariance. \\
\(\lambda_{\mathrm{int}},\rho_{\mathrm{pair}}\) &
Intervention-induced nuisance strength and paired-noise scale. \\
\(\rho_{\mathrm{err}},\epsilon_M\) & Bounds on
\(\|G-L_C\|_{\mathrm{op}}\) and
\(\|\bar G-G\|_{\mathrm{op}}\), respectively. \\
\bottomrule
\end{tabular}
\end{center}

\section{Experimental Details}\label{app:experiment-details}

This section provides the experimental details for the main results, including datasets, judge-output protocols, judge prompts, intervention specifications, and implementation details.

\subsection{Datasets}
\label{app:datasets}
We evaluate six continuous-score panels (Yelp, SummEval coherence, Summarize from Feedback, HelpSteer2, ASSET, Review-5K) and six binary/preference panels (Chatbot Arena, RewardBench 2, HelpSteer3, MT-Bench with human labels, LIAR2, CivilComments). In \cref{tab:continuous-real} the row labeled SummEval is the coherence aspect.

For continuous-score tasks, the target is the human-provided score and we report MAE. For binary/preference tasks, the target is the human-provided label or preference and we report accuracy.

\paragraph{Review-5K \citep{review5k}.}
Review-5K contains paper texts, reviewing guidelines, review records, and numeric reviewer ratings. We use the paper text and reviewing guidelines as the judge input; the reference review text is not shown to the judges. The target field is \texttt{avg\_rate}, the average available reviewer rating on the 1--10 review scale. The dataset is available as \href{https://huggingface.co/datasets/WestlakeNLP/Review-5K}{\texttt{WestlakeNLP/Review-5K}}.

\paragraph{Yelp \citep{yelp}.}
Yelp consists of user-written business reviews paired with star ratings. We use a 5{,}000-example subset and treat the task as continuous rating prediction: given the review text, the goal is to recover the associated 1--5 star rating. The source dataset is available as \href{https://huggingface.co/datasets/Yelp/yelp_review_full}{\texttt{Yelp/yelp\_review\_full}}.

\paragraph{ASSET \citep{asset_dataset}.}
ASSET is a sentence-simplification rating dataset. Each row contains a source sentence, a candidate simplification, an evaluation aspect, and a human rating. We use the \texttt{human\_rating} field as the continuous target on the 0--100 scale. We use 4{,}500 rating rows. The dataset is available as \href{https://huggingface.co/datasets/facebook/asset}{\texttt{facebook/asset}}.

\paragraph{HelpSteer2 \citep{wang2024helpsteer2}.}
HelpSteer2 contains prompts, model responses, and human ratings of five response attributes. We use the prompt and candidate response as the judge input and the helpfulness rating as the continuous target on the 0--4 scale, stored in the \texttt{score} field. We use the released validation split. The dataset is available as \href{https://huggingface.co/datasets/nvidia/HelpSteer2}{\texttt{nvidia/HelpSteer2}}.

\paragraph{Summarize from Feedback \citep{summarize_from_feedback_dataset}.}
Summarize from Feedback contains source posts, candidate summaries, and human assessments of summary quality. We use its \texttt{axis} subset, with the source post and candidate summary as the judge input. The target is the overall quality rating on the 1--7 scale, stored as \texttt{overall\_score} from \texttt{summary.axes.overall}. We pool the released validation and test splits before applying our evaluation protocol. The dataset is available as \href{https://huggingface.co/datasets/openai/summarize_from_feedback}{\texttt{openai/summarize\_from\_feedback}}.

\paragraph{SummEval \citep{fabbri2021summeval}.}
SummEval contains model-generated summaries of CNN/DailyMail news articles with human ratings of coherence, consistency, fluency, and relevance. We use the coherence panel: 1{,}600 summaries from 16 systems for 100 articles. The judge input is the article and candidate summary; the target is the mean expert coherence rating on the 1--5 scale, stored in the \texttt{label} field. The dataset is available from \href{https://github.com/Yale-LILY/SummEval}{\texttt{Yale-LILY/SummEval}}.

\paragraph{Chatbot Arena \citep{zheng2023judging}.}
Chatbot Arena contains pairwise comparisons between two model responses to the same user prompt. We use it as a preference task: the target field is \texttt{gold\_label\_binary}, which encodes the human preference over candidates A and B. We use 5{,}000 sampled comparisons, excluding ties. The dataset is available as \href{https://huggingface.co/datasets/lmsys/chatbot_arena_conversations}{\texttt{lmsys/chatbot\_arena\_conversations}}.

\paragraph{CivilComments \citep{borkan2019civilcomments}.}
CivilComments consists of online comments annotated for toxicity. We use a 5{,}000-example subset, sampled equally from the two classes in the test split, as a binary toxicity-classification task, where the input is the comment text and the target indicates whether the comment is toxic. The binary target is obtained from the dataset's continuous \texttt{toxicity} score, with scores of at least 0.5 labeled toxic (1) and the rest non-toxic (0). The source dataset is available as \href{https://huggingface.co/datasets/google/civil_comments}{\texttt{google/civil\_comments}}.

\paragraph{LIAR2 \citep{xu2024enhanced}.}
LIAR2 contains short statements assessed by professional fact-checkers at PolitiFact. We use the 2{,}296 statements in the released test split, with only the statement text as the judge input. We collapse the six veracity classes into a binary target: pants-on-fire, false, and barely-true map to 0; half-true, mostly-true, and true map to 1. The target is stored in the \texttt{label} field. The dataset is available as \href{https://huggingface.co/datasets/chengxuphd/liar2}{\texttt{chengxuphd/liar2}}.

\paragraph{MT-Bench \citep{zheng2023judging}.}
MT-Bench provides human comparisons of model responses to questions spanning multiple task categories. We use the 3{,}355-comparison human panel from \href{https://huggingface.co/datasets/allenai/preference-test-sets}{\texttt{allenai/preference-test-sets}}, with the question and two candidate responses as the judge input. The target identifies the human-preferred response, adjusted for the displayed candidate order. The original human judgments are available as \href{https://huggingface.co/datasets/lmsys/mt_bench_human_judgments}{\texttt{lmsys/mt\_bench\_human\_judgments}}.

\paragraph{HelpSteer3 \citep{wang2025helpsteer3preference}.}
HelpSteer3 contains human preferences between model responses across code, general, multilingual, and STEM tasks. We sample 625 training examples per domain and remove tied preferences. The judge input is the last user message and the two candidate responses. The target is the direction of the \texttt{overall\_preference} rating, stored as \texttt{gold\_label\_binary}. We read this reference field from the judge files to preserve row alignment. The dataset is available as \href{https://huggingface.co/datasets/nvidia/HelpSteer3}{\texttt{nvidia/HelpSteer3}}.

\paragraph{RewardBench 2 \citep{malik2025rewardbench2}.}
RewardBench 2 tests response selection across factuality, instruction following, mathematics, safety, focus, and ties. We construct a pairwise panel from its 1{,}865 prompts by taking the first chosen and first rejected completion for each prompt. The judge input is the prompt and these two responses; the target identifies the chosen response in the displayed order. We align intervention rows by source-panel position because pair identifiers repeat within comparison groups. The dataset is available as \href{https://huggingface.co/datasets/allenai/reward-bench-2}{\texttt{allenai/reward-bench-2}}.

\subsection{Evaluation protocol}
\label{app:current-protocol}
\paragraph{Data splits and hyperparameter tuning.}
Following \citet{zhao2026care}, we use five random row-level splits per dataset, reserving 10\% for validation on continuous tasks and 15\% on binary/preference tasks. We select intervention families and hyperparameters using the validation set. For CARE, we follow the original paper's tuning procedure: CARE-SVD tunes its regularization, while CARE-Tensor tunes its regularization, rank, and posterior cutoff.

\paragraph{Candidate-pool construction.}
We used Codex and Fable to inspect dataset examples and propose potential nuisance attributes and corresponding intervention instructions. Pool construction was open-ended, with no prespecified stopping rule for expansion. The pool was refined using experimental feedback, including validation metrics. Given a family specification, pair generation and preservation checks use no quality labels (\cref{appsec:intervention-prompts}).

\paragraph{Family selection.}
We select families by forward greedy search using mean validation performance across seeds. We first select the best-performing single family. At each subsequent step, we evaluate adding each remaining family, retuning \((\alpha,\gamma)\) for every candidate set, and accept the addition with the largest validation improvement. The search stops when no addition improves validation performance or 15 families have been selected.

\paragraph{Estimator settings.}
We tune \(\alpha\in\{0,0.1,0.25,0.5,0.75,1\}\) and \(\gamma\in\{0.25,1,4,16,64\}\), with \(\rho=\gamma\operatorname{tr}(\bar G)/p\). We select one configuration per dataset by minimizing mean validation MAE for continuous tasks or maximizing mean validation accuracy for binary/preference tasks across the five splits. Table~\ref{tab:configuration-variability} reports the selected configurations, family counts, and evaluation variability.
\begin{table}[t]
\centering\small
\setlength{\tabcolsep}{3pt}
\begin{tabular}{@{}lrrcc@{}}
\toprule
Dataset & Best baseline & \voirdire{} & \shortstack{Selected / candidate\\families} & $(\alpha,\gamma)$ \\
\midrule
\multicolumn{5}{@{}l}{\textit{Continuous scoring: MAE ($\downarrow$)}} \\
Yelp & $0.6940 \pm 0.0040$ & $0.5080 \pm 0.0156$ & 4 / 30 & (1, 64) \\
SummEval & $0.7293 \pm 0.0048$ & $0.7012 \pm 0.0052$ & 4 / 24 & (0, 1) \\
Summarize from Feedback & $1.3922 \pm 0.0136$ & $1.3461 \pm 0.0036$ & 4 / 41 & (0, 0.25) \\
HelpSteer2 & $0.9814 \pm 0.0057$ & $0.9514 \pm 0.0126$ & 3 / 33 & (0.75, 0.25) \\
ASSET & $27.6290 \pm 0.1560$ & $26.9303 \pm 0.3421$ & 5 / 53 & (0.5, 0.25) \\
Review-5K & $1.9570 \pm 0.0180$ & $1.8624 \pm 0.0352$ & 4 / 25 & (0.25, 64) \\
\midrule
\multicolumn{5}{@{}l}{\textit{Binary/preference: accuracy ($\uparrow$)}} \\
Chatbot Arena & $0.5959 \pm 0.0162$ & $0.6465 \pm 0.0105$ & 5 / 43 & (0.5, 0.25) \\
RewardBench 2 & $0.6052 \pm 0.0116$ & $0.6579 \pm 0.0065$ & 4 / 37 & (1, 4) \\
LIAR2 & $0.6780 \pm 0.0039$ & $0.6857 \pm 0.0047$ & 2 / 12 & (0.1, 0.25) \\
HelpSteer3 & $0.5921 \pm 0.0272$ & $0.6101 \pm 0.0146$ & 6 / 39 & (0.5, 0.25) \\
CivilComments & $0.7473 \pm 0.0031$ & $0.7387 \pm 0.0033$ & 2 / 22 & (0, 0.25) \\
MT-Bench (human) & $0.6128 \pm 0.0036$ & $0.6242 \pm 0.0065$ & 4 / 38 & (0.1, 1) \\
\bottomrule
\end{tabular}
\caption{Selected configurations and mean $\pm$ SD. Best baseline is the strongest mean among MV, AVG, ISP, PCA, and the available CARE variants.}
\label{tab:configuration-variability}
\end{table}

\paragraph{Weight orientation and normalization.}
The generalized eigenvector is defined only up to sign and scale. We orient the weights to have a positive sum and rescale them to sum to one, so the aggregate stays on the judges' score scale.

\paragraph{Computational resources and cost.}
We ran experiments on NVIDIA A100 40GB and L40 GPUs. Generating the original judge outputs took up to approximately three hours per dataset. Judge scoring and rewriting use GPUs, while estimator fitting and family selection run on CPUs. For \(k\) selected families and \(p\) judges, calibration uses \(32kp\) rewritten scores. Total cost also includes generating and scoring candidate interventions that are not selected.

\subsection{LLM Judge Models}
\label{appsec:judge-models}

Following \citet{zhao2026care}, we use fixed pools of open-weight LLM judges and filter their outputs for valid scores. For free-form scalar outputs, we retain judges with valid scores on more than 70\% of items, exclude items with missing, unparsable, or out-of-range scores from any retained judge, and remove judges with constant scores. The resulting panels contain 11 judges for continuous scoring and 20 for binary classification and pairwise preference.

\paragraph{Continuous-score judges.}
The continuous-score pool contains \texttt{Qwen3-\{0.6,1.7,4,8\}B}~\citep{qwen3}, \texttt{Llama-3.1-8B-Instruct} and \texttt{Llama-3.2-\{1,3\}B}~\citep{grattafiori2024llama}, \texttt{Gemma-3-\{1,4\}B-IT}~\citep{team2025gemma}, \texttt{Phi-4-mini-instruct}~\citep{abouelenin2025phi}, and \texttt{Mistral-7B-Instruct-v0.3}~\citep{jiang2023mistral7b}. Judges are prompted to return a scalar score on the dataset-native range: 1--10 for Review-5K, 1--5 for Yelp, 1--5 for SummEval coherence, 1--7 for Summarize from Feedback, 0--4 for HelpSteer2, and 0--100 for ASSET.

\paragraph{Binary and preference judges.}
The binary/preference pool is the same 20 judges on every panel: Qwen2.5 Instruct (1.5B, 3B, 7B, and 14B)~\citep{qwen25} and \texttt{Qwen3-\{0.6,1.7,4,8\}B}~\citep{qwen3}, \texttt{Llama-3.1-8B-Instruct}, \texttt{Llama-3.2-\{1,3\}B}, and \texttt{Llama-3.2-\{1,3\}B-Instruct}~\citep{grattafiori2024llama}, \texttt{Mistral-7B-Instruct-v0.3}~\citep{jiang2023mistral7b}, \texttt{Phi-4-mini-instruct}~\citep{abouelenin2025phi}, \texttt{Yi-1.5-\{6,9\}B-Chat}~\citep{young2024yi}, \texttt{Gemma-2-9B-IT}~\citep{team2024gemma2}, and \texttt{Gemma-3-\{1,4\}B-IT}~\citep{team2025gemma}. We map CivilComments and LIAR2 outputs to binary labels and pairwise outputs to candidate preferences for accuracy evaluation. The judge pool remains fixed throughout intervention-family selection.

\subsection{LLM Judge Prompt Templates}
\label{appsec:judge-prompts}

\paragraph{Prompting and score parsing.}
We use task-specific prompts and score ranges, applying the same prompt format and parsing rules to original and modified inputs. Ratings are extracted from the requested verdict field; pairwise ratings are converted to signed preferences and evaluated using a fixed threshold. Missing, unparsable, or out-of-range ratings are discarded. All judges receive raw completion prompts without chat templates. The templates below use placeholders in braces for item-specific content.

\paragraph{Review-5K paper scoring.}
The \texttt{\{guidelines\}} field holds the venue's reviewer guidelines and \texttt{\{paper\}} the submission text.
\begin{lstlisting}[style=prompt]
You are scoring the overall quality of a research paper submission. Use the reviewing guidelines below to assess methodological soundness, clarity, contribution, and overall merit. Assign one of the standard conference rating levels (integer in [1, 10]) using this rubric:

1 - Strong Reject: fundamental flaws; clearly unsuitable for publication.
3 - Reject, not good enough: major weaknesses outweigh contributions.
5 - Marginally below the acceptance threshold: borderline but leaning negative.
6 - Marginally above the acceptance threshold: borderline but leaning positive; improvements still desirable.
8 - Accept, good paper: solid contribution with meaningful advances and only minor issues.
10 - Strong Accept, should be highlighted at the conference: outstanding paper with exceptional clarity and impact.

If you believe an intermediate score (2, 4, 7, 9) better reflects the work, you may use it, positioning it between the adjacent rubric descriptions above.

Reviewing guidelines:
{guidelines}

Respond strictly in this format:

Feedback:::
Total rating: X

Where X is an integer between 1 and 10 inclusive. Output nothing else.

Paper content:
{paper}

Feedback:::
Total rating: 
\end{lstlisting}

\paragraph{Yelp review scoring.}
\leavevmode\par\smallskip
\begin{lstlisting}[style=prompt]
You are evaluating the overall sentiment of a Yelp restaurant review. Read the entire review, consider the reviewer's tone, described experiences, and final impression, then assign a single star rating from 1 to 5 according to this rubric:

1 - Strongly negative; serious problems or clear dissatisfaction.
2 - Mostly negative; several complaints outweigh positives.
3 - Mixed or neutral; balanced feedback with no strong lean.
4 - Mostly positive; satisfied with only minor issues.
5 - Highly positive; enthusiastic praise and strong recommendation.

Base the rating on the reviewer's intent, not your personal opinion, and ignore formatting quirks.

Respond exactly in this format:

Feedback:::
Total rating: X

Where X is an integer from 1 to 5 and nothing else.

Review:
{text}

Feedback:::
Total rating: 
\end{lstlisting}

\paragraph{SummEval coherence scoring.}
\leavevmode\par\smallskip
\begin{lstlisting}[style=prompt]
You are rating how coherent a candidate summary is. Read the article and the summary, judge whether the summary is well structured and its sentences follow one another in a sensible order, rather than reading as a disconnected list of facts, and output a single rating in [1, 5]:

1 - Disjointed; sentences do not connect and the order is arbitrary.
2 - Mostly disconnected, with abrupt transitions between sentences.
3 - Partly organised; some sentences follow naturally and others do not.
4 - Well organised, with only minor lapses in flow.
5 - Fully coherent; the summary reads as a single connected piece of text.

Judge organisation and flow only. Do not reward a summary for covering more of the article.

Respond strictly in this format:

Feedback:::
Total rating: X

Where X is an integer 1-5. Output nothing else.

Article:
{article}

Candidate summary:
{summary}

Feedback:::
Total rating: 
\end{lstlisting}

\paragraph{Summarize from Feedback scoring.}
\leavevmode\par\smallskip
\begin{lstlisting}[style=prompt]
You are an expert editor judging how faithfully a candidate summary captures the key information in the original Reddit post.
Read the post first (central events, reasons, outcomes). Then read the summary and assess faithfulness and coverage.

Score substance over surface:
- Prioritise factual faithfulness and coverage of core facts (who/what/when/where/why/how) and outcomes.
- Prefer precise, specific content grounded in the post; do not reward length, flourish, or formatting.
- Subtract for invented facts, contradictions, or missing main points; minor wording differences should rarely affect the score.
- A concise but accurate and comprehensive summary can score high; a longer but vague or extraneous one can score low.

Use the absolute 1-7 scale (not relative to other items):
1 - Completely inaccurate or misleading; major facts are wrong or missing.
2 - Very inaccurate; severe omissions or contradictions; little usable content.
3 - Poor; captures a few ideas but misses several important points or introduces errors.
4 - Fair; mostly correct but notable gaps, extraneous details, or minor hallucinations.
5 - Good; covers main ideas with small omissions or wording issues.
6 - Very good; accurate and fluent with only trivial gaps.
7 - Excellent; fully faithful, comprehensive, and well-written.

Guidance:
- Reserve 7 for both faithful and comprehensive summaries; use 1-2 for clearly poor ones.
- Minor style differences are acceptable if the content remains faithful.

Respond exactly in this format:

Feedback:::
Total rating: X

Where X is exactly one of: 1, 2, 3, 4, 5, 6, 7. Output nothing else.

Post:
{post}

Summary:
{summary}

Feedback:::
Total rating: 
\end{lstlisting}

\paragraph{HelpSteer2 response scoring.}
\leavevmode\par\smallskip
\begin{lstlisting}[style=prompt]
You are scoring how well the system_answer addresses the user_question.
Return a single float in [0.0, 4.0] (decimals allowed). Use the full range:
0.0 = Unhelpful or harmful; refusal; mostly wrong or off-topic.
1.0 = Barely helpful; major gaps or serious inaccuracies.
2.0 = Partially helpful; mixed quality with notable omissions or minor errors.
3.0 = Helpful and mostly correct; small omissions or minor issues only.
4.0 = Fully correct, helpful, and clear.

Rules:
- Prioritise factual accuracy, relevance, and completeness; do not reward verbosity.
- Penalize hallucinations, unsafe advice, or refusal when an answer is expected.
- If your internal judgment falls outside [0, 4], output the nearest bound instead.

Respond exactly in this format and nothing else:

Feedback:::
Total rating: X

Question: {question}
Answer: {answer}

Feedback:::
Total rating: 
\end{lstlisting}

\paragraph{ASSET simplification scoring.}
The \texttt{\{aspect\}} field is the row's rated aspect (meaning, fluency, or simplicity).
\begin{lstlisting}[style=prompt]
You are rating the quality of a simplified sentence for a specific aspect of text simplification.
Examine both sentences carefully before assigning a single integer score in [0, 100]. Use the whole range; do not default to 50 or round to multiples of 5 unless warranted.

Aspect to evaluate: {aspect}

Interpret the aspect as follows (case-insensitive):
- meaning / meaning preservation -- Judge how faithfully the simplification preserves the original meaning (facts, relations, intent). Ignore simplicity/fluency except where they alter meaning.
- fluency / grammaticality -- Judge grammatical correctness and naturalness of the simplification. Ignore meaning differences and degree of simplification; do not reward added content.
- simplicity / readability -- Judge how much easier to read the simplification is compared with the original (shorter sentences, simpler vocabulary/structure). Minor meaning loss may be acceptable, but removing essential information or making text ambiguous should not receive top scores.
- other -- If the aspect is unrecognised, judge overall simplification quality, prioritising meaning preservation, then fluency, then simplicity.

Anchor guidance (choose any integer within each band):
0-20  Unusable for the aspect.
21-40 Poor for the aspect.
41-60 Mixed: noticeable problems but some positives.
61-80 Good: solid performance with minor issues.
81-95 Very good: strong performance with only tiny flaws.
96-100 Excellent for the aspect.

General rules:
- Focus strictly on the stated aspect when scoring.
- Do not reward hallucinations or added facts not in the original.
- Non-round integers are allowed (e.g., 73, 88) if they best reflect quality.

Respond strictly in this format:

Feedback:::
Total rating: X

Where X is an integer in [0, 100]. Output nothing else.

Original sentence:
{original}

Candidate simplification:
{simplification}

Feedback:::
Total rating: 
\end{lstlisting}

\paragraph{CivilComments toxicity scoring.}
The 0--9 score is thresholded at 4.5 to obtain the binary toxicity decision.
\begin{lstlisting}[style=prompt]
You are a toxicity rater. Given a single user comment, assign a discrete toxicity score on a 0-9 scale. Higher numbers indicate more toxic language.

Return exactly one integer using this rubric:
0 : Clearly non-toxic or benign language.
1 : Mostly neutral with only faint negativity.
2 : Slightly negative or dismissive language.
3 : Mild insults or disrespectful tone.
4 : Noticeable harassment, repeated negativity, or coarse language.
5 : Borderline toxic; sustained insults or hostility.
6 : Clearly toxic with direct insults or slurs.
7 : Highly toxic; explicit harassment or hateful language.
8 : Severe toxicity including threats or strong hate speech.
9 : Extremely toxic content with explicit threats or violent hate speech.

Guidelines:
- Focus on the literal wording of the comment.
- Penalise hate, harassment, threats, or extreme vulgarity.
- If the comment mixes toxic and mitigating language, choose the score that reflects the overall impact.

Respond strictly in this format:

Feedback:::
Total rating: X

Where X is an integer between 0 and 9 (inclusive). Output nothing else.

Comment: {text}

Feedback:::
Total rating: 
\end{lstlisting}

\paragraph{LIAR2 veracity scoring.}
The 0--9 score is thresholded at 4.5 to obtain the binary truthfulness decision.
\begin{lstlisting}[style=prompt]
You are a fact-checking rater. Given a single political claim, assign a discrete accuracy score on a 0-9 scale. Higher numbers indicate a claim that is more accurate.

Return exactly one integer using this rubric:
0 : Fabricated; no factual basis at all.
1 : Clearly false and contradicted by available evidence.
2 : False; the central assertion is not supported.
3 : Mostly false; a kernel of truth badly distorted.
4 : Borderline false; the overall impression misleads.
5 : Mixed; as much misleading as accurate.
6 : Half true; a real basis but material omissions.
7 : Mostly true; minor details overstated.
8 : True with only trivial imprecision.
9 : Entirely true and well supported.

Guidelines:
- Judge the factual accuracy of the claim itself, not its tone or the speaker.
- Rhetoric, framing and political stance are not evidence.
- If a claim mixes accurate and inaccurate elements, score the overall impression it creates.

Respond strictly in this format:

Feedback:::
Total rating: X

Where X is one of: 0, 1, 2, 3, 4, 5, 6, 7, 8, 9. Output nothing else.

Statement: {text}

Feedback:::
Total rating: 
\end{lstlisting}

\paragraph{Pairwise preference scoring.}
Ratings below 3 favor response A, while ratings above 3 favor response B. HelpSteer3, RewardBench~2, and MT-Bench use the first template, which allows a tie at 3. Chatbot Arena uses the second template, which excludes ties.
\begin{lstlisting}[style=prompt]
You will compare two assistant responses (A and B) to the same user question. Judge which response is more helpful overall, balancing factual accuracy, completeness, clarity, safety, and tone.

Return a single integer chosen from {0, 1, 2, 3, 4, 5, 6} using this rubric:
0 : Response A is far superior to Response B
1 : Response A is clearly better than Response B
2 : Response A is slightly better than Response B
3 : Responses are essentially tied (equally good or bad)
4 : Response B is slightly better than Response A
5 : Response B is clearly better than Response A
6 : Response B is far superior to Response A

Guidelines:
- Read the user question carefully to understand the user's goal.
- Evaluate each answer independently for correctness, helpfulness, and safety.
- If Response A is better overall, you must output one of {0, 1, 2}. If Response B is better overall, output one of {4, 5, 6}. If they are truly tied, output 3. Do not output any other number.
- Compare the answers and decide how strongly one is better using the rubric above.
- Penalize hallucinations, unsafe advice, or refusal when a substantive answer is expected.

Respond strictly in this format:

Feedback:::
Total rating: X

Where X is one of: 0, 1, 2, 3, 4, 5, 6. Output nothing else.

Question: {question}

Response A: {answer_a}

Response B: {answer_b}

Feedback:::
Total rating: 
\end{lstlisting}
\begin{lstlisting}[style=prompt]
You will compare two assistant responses (A and B) for the same user question. Judge which response better helps the user overall, considering factual accuracy, completeness, clarity, safety, and tone.

Return a single integer chosen from {0, 1, 2, 4, 5, 6} using this rubric:
0 : Response A is far superior to Response B
1 : Response A is clearly better than Response B
2 : Response A is slightly better than Response B
4 : Response B is slightly better than Response A
5 : Response B is clearly better than Response A
6 : Response B is far superior to Response A

Important (must follow all points below or the answer is invalid):
- Ties are NOT allowed. Never output 3 or any other number outside the set {0, 1, 2, 4, 5, 6}.
- If Response A is better overall, you MUST output one of {0, 1, 2}. Never output 4, 5, or 6 when Response A wins.
- If Response B is better overall, you MUST output one of {4, 5, 6}. Never output 0, 1, or 2 when Response B wins.
- If the responses are very close, choose 2 (if A is slightly better) or 4 (if B is slightly better).
- Output only the integer--no explanations, words, or additional text.

Remember: Response A better -> 0/1/2. Response B better -> 4/5/6. Outputs outside these sets will be rejected.

Respond strictly in this format (no extra text):

Feedback:::
Total rating: X

Where X is one of: 0, 1, 2, 4, 5, 6.

Question: {question}

Response A: {answer_a}

Response B: {answer_b}

Feedback:::
Total rating: 
\end{lstlisting}

\subsection{Intervention Specifications and Preservation Checks}
\label{appsec:intervention-prompts}

\paragraph{Intervention specifications.}
Each intervention family specifies (i) the nuisance attribute and input span to modify, (ii) the rule used to perform the modification, and (iii) the preservation constraints used to accept the resulting pair. Modification rules are either deterministic transformations or family-specific instructions executed by an LLM; in our experiments, we use Llama-3.1-8B-Instruct.

\paragraph{Rewrite prompts.}
The rewrite model receives the input text, a family-specific instruction, and preservation constraints. Word-overlap and length constraints are set according to the intervention family and input length. The model returns only the rewritten text. The following example shows the Yelp concision prompt.

\begin{lstlisting}[style=prompt]
You are rewriting text for a controlled intervention study.

Dataset: yelp_with_scores
Task type: continuous

Goal:
Express the same content more briefly. Remove redundancy and filler only. Keep every fact, number, name and claim. Do not add anything new.

Review:
I really like this place a lot better now that they've moved from the Borgata. It's a nice open floor plan and I love the outside patio and bar area. My pizza was great and I'm pleased to see their nice selection of Italian wines.

Hard constraints:
- Preserve the same meaning, main claims, recommendation, sentiment, safety posture, and overall quality judgment.
- Do not add, remove, contradict, or strengthen substantive content.
- Keep every number, date, name, and URL exactly as in the original.
- Keep every negation as in the original.
- Wording: reuse the original's own content words, so that at least 55 percent of them still appear in the rewrite.
- Length: the rewrite must be between 127 and 212 characters; the original is 230.

Return only the rewritten text.
\end{lstlisting}

\paragraph{Preservation checks.}
For deterministic transformations, checks verify the specified structural properties, such as preservation of text, letter sequences, or lines. Position swaps must exchange candidates verbatim; context and prompt modifications must leave the judged text unchanged.

LLM rewrites must preserve all numbers and the negation count, introduce no new URLs, and satisfy the content-word Jaccard similarity and length-ratio bounds in \cref{tab:intervention-specifications}. Families without a specified length range use \([0.70,1.60]\). These checks filter unintended changes but do not guarantee preservation of the evaluation target.

\paragraph{Intervention examples.}
\Cref{tab:yelp-selected-specs} lists the four families selected for Yelp and the position-swap intervention used on preference datasets. The examples below show their application to a Yelp review and a Chatbot Arena comparison. All pairs passed the preservation checks. The Yelp example is the shortest review shared by the four selected families' calibration pools, chosen without inspecting judge scores.

\begin{lstlisting}[style=prompt]
Original review:
I really like this place a lot better now that they've moved from the Borgata.  It's a nice open floor plan and I love the outside patio and bar area.  My pizza was great and I'm pleased to see their nice selection of Italian wines.

Concision (LLM rewrite):
I like this place more since they moved from the Borgata. The open floor plan and outside patio/bar area are nice. My pizza was great and they have a good Italian wine selection.

Prose (LLM rewrite):
I really like this place a lot better now that they've moved from the Borgata. The open floor plan is nice and I love the outside patio and bar area. My pizza was great and I'm pleased to see their nice selection of Italian wines.

Markdown quote (deterministic):
> I really like this place a lot better now that they've moved from the Borgata.  It's a nice open floor plan and I love the outside patio and bar area.  My pizza was great and I'm pleased to see their nice selection of Italian wines.

XML frame (deterministic):
<document>
I really like this place a lot better now that they've moved from the Borgata.  It's a nice open floor plan and I love the outside patio and bar area.  My pizza was great and I'm pleased to see their nice selection of Italian wines.
</document>
\end{lstlisting}

\begin{lstlisting}[style=prompt]
Shared user question:
Write me a haiku about hiroshima

Original order:
Response A: Hiroshima weeps,
Silent scars whisper of peace,
Healing through time's grace.
Response B: A flash of light,
A city in ruins,
A world changed forever.

After position swap:
Response A: A flash of light,
A city in ruins,
A world changed forever.
Response B: Hiroshima weeps,
Silent scars whisper of peace,
Healing through time's grace.
\end{lstlisting}

\begin{table*}[h]
\centering\small
\begin{tabular}{lrr}
\toprule
LLM axis & Minimum word Jaccard & Length ratio interval \\
\midrule
Presentation polish & 0.60 & [0.70, 1.60] \\
Concision & 0.55 & [0.55, 0.92] \\
Verbosity & 0.60 & [1.15, 2.00] \\
Authority register & 0.70 & [0.85, 1.50] \\
Conclusion framing & 0.80 & [0.95, 1.40] \\
Hedging & 0.70 & [0.95, 1.40] \\
Informality & 0.60 & [0.80, 1.30] \\
Prose form & 0.75 & [0.85, 1.20] \\
\bottomrule
\end{tabular}
\caption{Family-specific LLM rewrite checks, in addition to the common invariants.}
\label{tab:intervention-specifications}
\end{table*}

\begin{table*}[h]
\centering\small
\setlength{\tabcolsep}{4pt}
\begin{tabular}{llp{6.6cm}l}
\toprule
Family & Kind & Operation & Preservation check \\
\midrule
Concision & LLM & ``Express the same content more briefly. Remove redundancy and filler only. Keep every fact, number, name and claim. Do not add anything new.'' & Jaccard $\ge 0.55$, length $[0.55, 0.92]$ \\
Prose form & LLM & ``Convert any headings, bullet points, numbering or emphasis into continuous prose, keeping the sentences in their original order and wording. Do not summarize, omit or add anything.'' & Jaccard $\ge 0.75$, length $[0.85, 1.20]$ \\
Markdown quote & deterministic & Prefix every line of the exact text with ``\texttt{> }''. & letter sequence equal \\
XML frame & deterministic & Wrap the exact text in \texttt{<document>} tags. & exact text preserved \\
Position swap & deterministic & Exchange the two candidates verbatim; the prompt-order score is negated at fit time. & exact candidate exchange \\
\bottomrule
\end{tabular}
\caption{The four families selected for Yelp and the position-swap family of the preference panels. All LLM families additionally pass the common checks of \cref{appsec:intervention-prompts}.}
\label{tab:yelp-selected-specs}
\end{table*}

\paragraph{Family availability.}
Only families with enough valid, fully scored pairs enter selection, yielding 12--53 eligible families per dataset. On Review-5K, LLM rewrites failed the number-preservation checks, so calibration uses only deterministic structural transformations.

\subsection{Controlled Judge-Bias Benchmark}
\label{app:injected-bias}

We describe the benchmark and protocol used in \cref{subsec:injected-bias}.

\paragraph{Data and judges.}
The benchmark of \citet{chen-etal-2024-humans} contains two GPT-4 answers to each of 142 questions. We reuse its GPT-4 rewrites targeting rich formatting (beauty) or added references (authority); these can also rephrase or expand the answer. We use the scores provided by CARE \citep{zhao2026care}, following its preprocessing to exclude Phi-4-mini and retain pairs with valid original and modified scores. The resulting panels contain ten judges rating answers on a 1--10 scale, with 136--142 valid pairs per modification. \Cref{fig:injected-bias-examples} illustrates the beauty and authority rewrites.

\begin{figure}[!htbp]
    \centering\small\setlength{\fboxsep}{4pt}
    \begin{minipage}[t]{0.32\linewidth}
    \colorbox{gray!8}{\parbox[t]{\dimexpr\linewidth-2\fboxsep}{%
    \textbf{Original}\\
    \textit{Why do we have night and day?}\\
    Night and day occur due to the Earth's rotation on its axis. When a part of the Earth is facing the Sun, it experiences daylight. [\ldots]}}
    \end{minipage}\hfill
    \begin{minipage}[t]{0.32\linewidth}
    \colorbox{orange!12}{\parbox[t]{\dimexpr\linewidth-2\fboxsep}{%
    \textbf{Beauty}\\
    emojis and Markdown\\
    \# \earth{} Earth's Rotation: Night \& Day \sun\leftmoon\\
    Night and day occur due to the **Earth's rotation** on its axis. When a part of the Earth is facing the Sun \sun, it experiences **daylight**. [\ldots]}}
    \end{minipage}\hfill
    \begin{minipage}[t]{0.32\linewidth}
    \colorbox{blue!8}{\parbox[t]{\dimexpr\linewidth-2\fboxsep}{%
    \textbf{Authority}\\
    fake citation\\
    {[\ldots]} This rotation happens over a period of 24 hours, creating our cycle of day and night \textbf{(National Geographic, 2019)}.}}
    \end{minipage}
    \caption{One benchmark answer and its beauty and authority rewrites \citep{chen-etal-2024-humans}. Emojis are rendered as symbols; the fake citation is in bold.}
    \label{fig:injected-bias-examples}
\end{figure}

\paragraph{Calibration and evaluation.}
We use beauty and authority pairs as calibration interventions. For each bias, we partition the pairs into five folds and repeat the procedure over ten seeds. The main comparison uses 32 calibration pairs sampled outside the evaluation fold. All methods use the same evaluation folds. We measure the mean absolute change in aggregate scores between original and modified answers.

PCA and CARE-SVD are fitted to the full original-score panel and applied to both versions. \voirdire{} uses the same original panel together with the sampled calibration pairs. ISP is computed jointly on original and modified scores. Evaluation items therefore contribute unlabeled scores to fitting, but their intervention pairs are excluded from calibration.

\paragraph{Parameter settings.}
We fix \((\alpha,\gamma)=(1,1/4)\) across biases, folds, seeds, and calibration budgets. We use maximum attenuation, \(\alpha=1\), to suppress the known nuisance manipulation. The value \(\gamma=1/4\) is the most frequently selected setting on the six continuous-score benchmarks, selected on three of them, giving \(\rho=\operatorname{tr}(\bar G)/(4p)\). No quality labels from this benchmark are used for fitting or parameter selection.

\FloatBarrier
\subsection{Intervention-Pair Budget: Details and Additional Results}
\label{app:pairs-per-family-budget}

We provide sampling details and per-dataset results for \cref{subsec:pairs-ablation}.

\paragraph{Setup.}
We vary the number of pairs per family over \(m\in\{4,8,16,32,64\}\), keeping the selected intervention families fixed. Samples are nested within each seed: \(m=32\) reproduces the main experiments, smaller budgets use subsets, and \(m=64\) adds available pairs. We re-select \((\alpha,\gamma)\) on validation at each budget, as in the main-body ablation. All budgets use the same five splits and evaluation items. We report means and one standard error.

\begin{figure*}[!t]
\centering
\includegraphics[width=\textwidth]{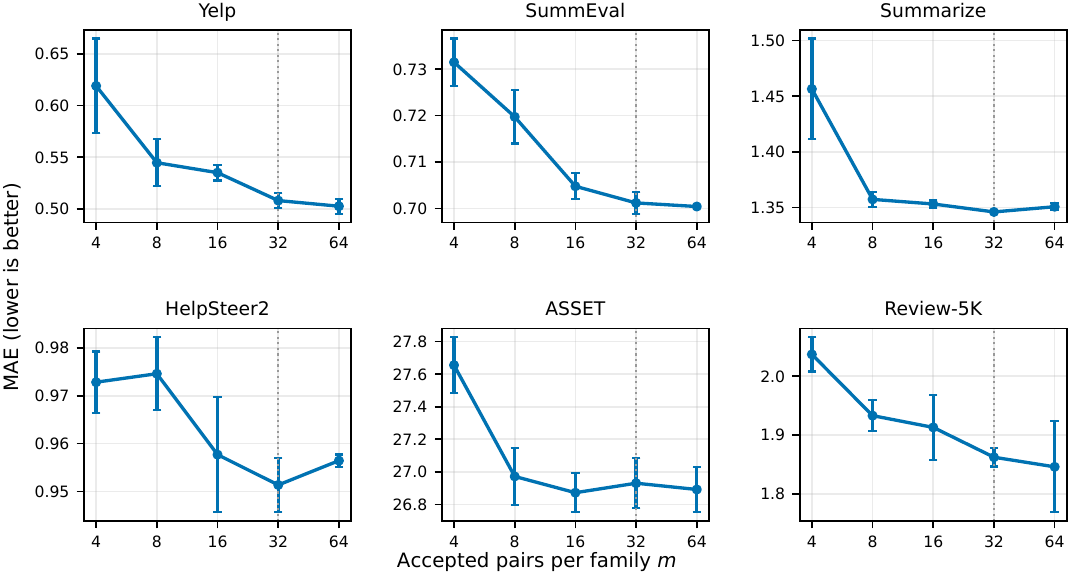}
\caption{Per-dataset MAE as the number of intervention pairs per family varies; lower is better. Error bars show one standard error across five splits. The dotted line marks the main experimental budget, \(m=32\).}
\label{fig:count-ablation-continuous}
\end{figure*}

\begin{figure*}[!t]
\centering
\includegraphics[width=\textwidth]{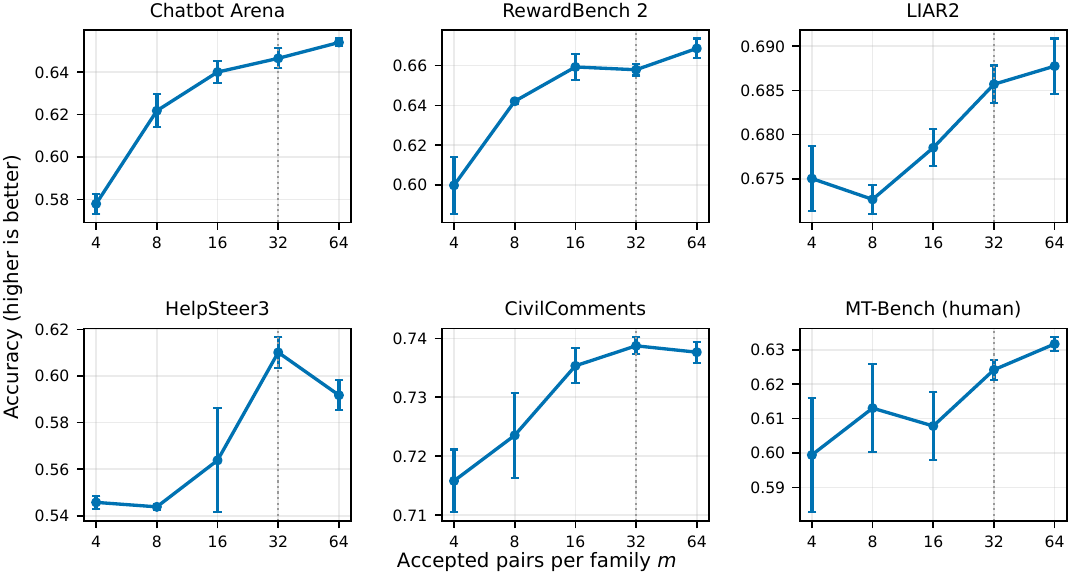}
\caption{Per-dataset accuracy as the number of intervention pairs per family varies; higher is better. Error bars show one standard error across five splits. The dotted line marks the main experimental budget, \(m=32\).}
\label{fig:count-ablation-binary-preference}
\end{figure*}

\paragraph{Results.}
\Cref{fig:count-ablation-continuous,fig:count-ablation-binary-preference} show the per-dataset results summarized in \cref{fig:pairs-ablation}. Using 32 pairs outperforms both 4 and 8 pairs on all 12 datasets, while gains beyond 32 pairs are generally small and not uniform across datasets.

\FloatBarrier
\subsection{Related-Nuisance Transfer Details}
\label{app:chen-transfer-details}

We provide construction and evaluation details for \cref{subsec:intervention-transfer}, following the benchmark protocol in \cref{app:injected-bias}.

\paragraph{Intervention construction.}
We use \texttt{Llama-3.1-8B-Instruct} to generate numbered structure, minimalist panels, and topic-specific icons. Prompts request presentation changes while preserving the original wording and order. We use temperature zero and a 1,024-token limit, retaining the first completion without preservation filtering or retries. Expert attribution, a source header, and fabricated footnotes use deterministic insertions. The source header attributes the answer to a generic reference handbook, without adding an answer-specific citation as in the benchmark authority rewrite. \Cref{tab:chen-transfer-edits} lists the specifications.

\begin{table}[!htbp]
    \centering\small
    \begin{tabular}{@{}lp{0.69\linewidth}@{}}
    \toprule
    Variant & Requested rewrite or deterministic operation \\
    \midrule
    \multicolumn{2}{@{}l}{\emph{Beauty: LLM-generated formatting}} \\
    Numbered structure & \raggedright Number existing steps or parallel points
    consecutively as \texttt{1.}, \texttt{2.}, etc. Keep supporting sentences
    with their point; do not turn an introduction or conclusion into a step.
    Preserve words such as ``First'' and ``Finally.'' Add no headings,
    icons, bolding, or blockquotes. \tabularnewline
    Minimalist panels & \raggedright Separate existing logical blocks with
    \texttt{-{}-{}-} on its own line. Mark one existing complete answer or
    conclusion sentence with \texttt{>} in its original position, without
    duplication or rewording. Add no headings, emojis, bolding, or lists.
    \tabularnewline
    Topic-specific icons & \raggedright Prefix content blocks with at most one
    topic-relevant icon from the fixed neutral palette. Preserve the original
    text and add no other formatting. \tabularnewline
    \midrule
    \multicolumn{2}{@{}l}{\emph{Authority: deterministic insertions}} \\
    Expert & \raggedright Prepend \texttt{Author: a PhD-qualified specialist in this subject.}
    followed by two newline characters. \tabularnewline
    Source header & \raggedright Prepend \texttt{Source: a university reference handbook.}
    followed by two newline characters. \tabularnewline
    Footnote & \raggedright Insert \texttt{[1]} immediately after the first detected
    sentence. Append two newline characters and
    \texttt{[1] Academic Reference Board. Reference Handbook. University Press, 2024.} \tabularnewline
    \bottomrule
    \end{tabular}
    \caption{Calibration variants in \cref{tab:intervention-transfer}. Numbered, Panels, and Icons abbreviate the beauty variants in the result tables. Beauty outputs are not filtered for compliance with these instructions. All authority credentials, source attributions, and reference metadata are fabricated experimental cues.}
    \label{tab:chen-transfer-edits}
\end{table}

Examples of the resulting rewrites appear below.

\begin{lstlisting}[style=prompt]
Question: What is the freezing point of water in Fahrenheit?

Original:
The freezing point of water in Fahrenheit is 32 degrees. This is based on the Fahrenheit scale, where the freezing and boiling points of water are set 180 degrees apart, with the freezing point at 32 and the boiling point at 212.

Numbered structure:
1. The freezing point of water in Fahrenheit is 32 degrees. 
2. This is based on the Fahrenheit scale, where the freezing and boiling points of water are set 180 degrees apart, with the freezing point at 32 and the boiling point at 212.

Minimalist panels:
--- 
The freezing point of water in Fahrenheit is 32 degrees. 
This is based on the Fahrenheit scale, where the freezing and boiling points of water are set 180 degrees apart, with the freezing point at 32 and the boiling point at 212.

Topic-specific icons:
[books] The freezing point of water in Fahrenheit is 32 degrees. This is based on the Fahrenheit scale, where the freezing and boiling points of water are set 180 degrees apart, with the freezing point at 32 and the boiling point at 212.

Expert:
Author: a PhD-qualified specialist in this subject.

The freezing point of water in Fahrenheit is 32 degrees. This is based on the Fahrenheit scale, where the freezing and boiling points of water are set 180 degrees apart, with the freezing point at 32 and the boiling point at 212.

Source header:
Source: a university reference handbook.

The freezing point of water in Fahrenheit is 32 degrees. This is based on the Fahrenheit scale, where the freezing and boiling points of water are set 180 degrees apart, with the freezing point at 32 and the boiling point at 212.

Footnote:
The freezing point of water in Fahrenheit is 32 degrees.[1] This is based on the Fahrenheit scale, where the freezing and boiling points of water are set 180 degrees apart, with the freezing point at 32 and the boiling point at 212.

[1] Academic Reference Board. Reference Handbook. University Press, 2024.
\end{lstlisting}

\paragraph{Calibration and evaluation.}
We use five folds and ten seeds with \((\alpha,\gamma)=(1,1/4)\). Each source family supplies 32 calibration pairs per split, excluding evaluation questions and requiring valid scores from all ten judges. Source shifts use newly scored original--rewrite pairs; target evaluation and covariance estimation reuse the benchmark scores from \cref{subsec:injected-bias}. Covariance estimation includes unlabeled original evaluation inputs. Each source--target model is frozen before evaluation.

\paragraph{Transfer metric.}
Let \(U_{a,b}\) denote the mean absolute score shift on target family \(b\) after calibration on source family \(a\), averaged over folds and seeds. We report the percentage of the target-family improvement over PCA retained by source calibration:
\[
R_{a,b}
=100\,
\frac{U_{\mathrm{PCA},b}-U_{a,b}}
     {U_{\mathrm{PCA},b}-U_{b,b}}.
\]
PCA corresponds to \(0\%\) and direct target-family calibration to \(100\%\). Values above \(100\%\) indicate lower sensitivity than the target-family reference; negative values indicate greater sensitivity than PCA.

\section{Additional Theory and Proofs}\label{appsec:theory-proofs}\label{app:theory-proofs}

This appendix provides the formal details for \cref{sec:theory}. We give full statements and proofs for observational non-identifiability and estimation error in the pooled nuisance matrix and its normalized attenuation operator, followed by five extensions:

\begin{itemize}[nosep,topsep=0pt,leftmargin=*]
    \item \textbf{Effect of imperfect interventions.} We quantify how violations of target preservation and unintended rewrite changes affect recovery of the nuisance-sensitive subspace.
    \item \textbf{Finite-sample estimation and recovery.} We show how the number of calibration pairs and the effective rank control moment-estimation error, and give the additional spectral-gap condition for nuisance-subspace recovery.
    \item \textbf{Effect of attenuation.} We show that attenuation preserves the target component and suppresses nuisance variation under an ideal orthogonal model, and bound target distortion due to paired judge noise and estimation error.
    \item \textbf{Perturbation of the aggregation weights.} We bound the leading generalized eigenvector up to relative sign when both matrices of its pencil change, and state the local conditions needed to transfer this bound through the score-normalization rule.
    \item \textbf{Multiple intervention families.} We justify the pooled second-moment estimator used by \voirdire{} when multiple selected intervention families move a common nuisance subspace.
\end{itemize}

\subsection{Assumptions}\label{app:graphical-model-assumptions}

We use the model in \cref{eq:minimal-intervention-model} to formalize when observational judge agreement is ambiguous. For an original input, the judge-output vector contains a target-bearing component \(U_Q q_i\) and a nuisance-sensitive component \(U_C c_i\). Observational aggregators only observe \(x_i^{(0)}\), so they must infer from unperturbed judge agreement which directions correspond to target variation. Without additional structure, this is not identifiable: target-relevant and nuisance-sensitive directions can be rotated without changing the observational distribution, as in standard latent-factor non-identifiability \citep{anderson1956statistical}.

The rewrites provide paired judge outputs for the same item before and after intervention. The key assumption is target invariance: the correct label, score, or preference remains fixed across the original and rewritten inputs, while the rewrite changes a nuisance attribute. In the ideal setting,
\[
    x_i^{(0)} = U_Q q_i + U_C c_i + \varepsilon_i^{(0)},\]
\[
    x_i^{(a)} = U_Q q_i + U_C(c_i+\delta_i^{(a)})
    + \varepsilon_i^{(a)} .
\]
Thus the target factor \(q_i\) is shared by the two judge-output vectors, while \(\delta_i^{(a)}\) captures the nuisance movement induced by intervention family \(a\).

The analysis does not require judge errors to be independent. Correlated judge noise is allowed and enters through the paired residual \(\xi_i^{(a)}=\varepsilon_i^{(a)}-\varepsilon_i^{(0)}\). What matters is the paired-difference contrast:
\[
    z_i^{(a)}
    :=
    x_i^{(a)}-x_i^{(0)}
    =
    U_C\delta_i^{(a)}+\xi_i^{(a)} .
\]
Because the target term is common to the original and rewritten inputs, it cancels in this difference. We assume \(\mathbb E[\xi_i^{(a)}]=0\) and that \(\xi_i^{(a)}\) is independent of \(\delta_i^{(a)}\); the nuisance movement \(\delta_i^{(a)}\) may have a nonzero mean. The paired differences therefore identify directions that respond to the intervention. \voirdire{} then attenuates these nuisance-sensitive directions in the original judge outputs before aggregation.

\subsection{Observational non-identifiability: full statement and proof}
\label{app:observational-full}

\begin{fullstatement}{Theorem~\ref{thm:observational-nonidentifiability}}
Consider the noiseless one-target, one-nuisance model
\[
    x_i^{(0)} = u q_i + v c_i,
\]
where \(u,v\in\mathbb R^p\) are orthonormal, \((q_i,c_i)\sim N(0,I_2)\), and \(q_i\) is the target-bearing coordinate. For any \(\theta\in(0,\pi/2)\), there exist rotated directions \(u_\theta,v_\theta\) and rotated coordinates \(q_{i,\theta},c_{i,\theta}\) such that
\[
    x_i^{(0)}
    =
    u_\theta q_{i,\theta}+v_\theta c_{i,\theta},
    \qquad
    (q_{i,\theta},c_{i,\theta})\sim N(0,I_2).
\]
Thus the original and rotated representations induce the same observational distribution over judge outputs, but assign different directions to the target-bearing coordinate. Therefore, the target direction is not identifiable from unperturbed judge outputs alone.
\end{fullstatement}

\paragraph{Proof of \cref{thm:observational-nonidentifiability}.}
Let
\[
    R_\theta =
    \begin{bmatrix}
        \cos\theta & -\sin\theta\\
        \sin\theta & \cos\theta
    \end{bmatrix}.
\]
Define the rotated directions and coordinates by
\[
    [u_\theta\ v_\theta] = [u\ v]R_\theta,
    \qquad
    \begin{bmatrix}
        q_{i,\theta}\\
        c_{i,\theta}
    \end{bmatrix}
    =
    R_\theta^\top
    \begin{bmatrix}
        q_i\\
        c_i
    \end{bmatrix}.
\]
Since \(R_\theta\) is orthogonal, \((q_{i,\theta},c_{i,\theta})\sim N(0,I_2)\). Moreover,
\[
\begin{aligned}
    u_\theta q_{i,\theta}+v_\theta c_{i,\theta}
    &=
    [u\ v]R_\theta R_\theta^\top
    \begin{bmatrix}
        q_i\\
        c_i
    \end{bmatrix}  \\
    &=
    u q_i + v c_i
    =
    x_i^{(0)} .
\end{aligned}
\]
Therefore the original and rotated parameterizations generate exactly the same marginal distribution over the observed judge-output vector \(x_i^{(0)}\).

However, they assign the target-bearing coordinate to different directions: the original representation uses \(u\), while the rotated representation uses \(u_\theta\). Since \(u_\theta\neq u\) for \(\theta\in(0,\pi/2)\), these are distinct target directions. Thus the same observational distribution is compatible with different choices of the target-bearing direction, so the target direction is not identifiable from unperturbed judge outputs alone.
\qed

\subsection{Nuisance-matrix estimation: full statement and proof}
\label{app:paired-difference-full}

The estimator uses the entire pooled second moment in both attenuation and the penalty in intervention-aware aggregation. We first control this matrix and its normalization, without requiring a nuisance rank or an eigengap.

\begin{fullstatement}{Proposition~\ref{prop:paired-difference}}
For fixed family weights \(\pi_a\ge0\) with \(\sum_a\pi_a=1\), define
\[
\begin{aligned}
    A_a&:=\mathbb E[\delta_i^{(a)}\delta_i^{(a)\top}]
        =\Cov(\delta_i^{(a)})+\mu_a\mu_a^\top,
        \qquad \mu_a:=\mathbb E[\delta_i^{(a)}],\\
    L_C&:=U_C\left(\sum_a\pi_a A_a\right)U_C^\top,
        \qquad
    G:=\sum_a\pi_a\,\mathbb E[z_i^{(a)}z_i^{(a)\top}].
\end{aligned}
\]
The pair-weighted estimator uses \(\pi_a=m_a/M\). Equal family budgets \(m_a=m\) give \(M=km\) and \(\pi_a=1/k\), as in the main text. Under \cref{eq:minimal-intervention-model}, centered paired noise independent of the nuisance movement gives
\begin{equation}\label{eq:paired-covariance-minimal}
    M_a:=\mathbb E[z_i^{(a)}z_i^{(a)\top}]
    =U_CA_aU_C^\top+\Omega_a,
    \qquad \Omega_a:=\Cov(\xi_i^{(a)}).
\end{equation}
Thus \(G=L_C+\sum_a\pi_a\Omega_a\) in the ideal intervention model.

More generally, write \(G=L_C+E_{\mathrm{err}}\), allowing paired noise and imperfect interventions, and suppose
\[
    \|E_{\mathrm{err}}\|_{\mathrm{op}}\le\rho_{\mathrm{err}},
    \qquad
    \|\bar G-G\|_{\mathrm{op}}\le\epsilon_M.
\]
Then the positive semidefinite empirical moment \(\bar G\) satisfies
\begin{equation}\label{eq:nuisance-matrix-error}
    \|\bar G-L_C\|_{\mathrm{op}}\le\rho_{\mathrm{err}}+\epsilon_M.
\end{equation}
If \(L_C\ne0\) and \(\bar G\ne0\), define
\[
    T:=\frac{L_C}{\lambda_{\max}(L_C)},
    \qquad
    \widehat T:=\frac{\bar G}{\lambda_{\max}(\bar G)}.
\]
Their difference obeys
\begin{equation}\label{eq:normalized-nuisance-error}
    \|\widehat T-T\|_{\mathrm{op}}
    \le\frac{2(\rho_{\mathrm{err}}+\epsilon_M)}{\lambda_{\max}(L_C)}.
\end{equation}
These deterministic bounds require neither independent sampled rows nor a positive semidefinite perturbation \(E_{\mathrm{err}}\).
\end{fullstatement}

\begin{proof}[Proof of \cref{prop:paired-difference}]
Subtracting the two views gives \(z_i^{(a)}=U_C\delta_i^{(a)}+\xi_i^{(a)}\). The cross moment vanishes because
\[
    \mathbb E[U_C\delta_i^{(a)}\xi_i^{(a)\top}]
    =U_C\,\mathbb E[\delta_i^{(a)}]\,
        \mathbb E[\xi_i^{(a)}]^\top=0.
\]
Expanding the second moment proves \cref{eq:paired-covariance-minimal}; weighting by \(\pi_a\) yields the pooled decomposition. The identity \(A_a=\Cov(\delta_i^{(a)})+\mu_a\mu_a^\top\) retains systematic shifts.

For the general perturbation, the triangle inequality gives
\[
\begin{aligned}
    \|\bar G-L_C\|_{\mathrm{op}}
    &\le\|\bar G-G\|_{\mathrm{op}}+\|G-L_C\|_{\mathrm{op}}\\
    &\le\epsilon_M+\rho_{\mathrm{err}}.
\end{aligned}
\]
For nonzero \(L_C,\bar G\), put \(a=\|L_C\|_{\mathrm{op}}=\lambda_{\max}(L_C)>0\) and \(b=\|\bar G\|_{\mathrm{op}}=\lambda_{\max}(\bar G)>0\), using positive semidefiniteness. The decomposition
\[
    \frac{\bar G}{b}-\frac{L_C}{a}
    =\frac{\bar G-L_C}{a}
      +\left(\frac1b-\frac1a\right)\bar G
\]
and the reverse triangle inequality imply
\[
\begin{aligned}
    \|\widehat T-T\|_{\mathrm{op}}
    &\le\frac{\|\bar G-L_C\|_{\mathrm{op}}+|a-b|}{a}\\
    &\le\frac{2\|\bar G-L_C\|_{\mathrm{op}}}{a}
    \le\frac{2(\rho_{\mathrm{err}}+\epsilon_M)}{\lambda_{\max}(L_C)}.
\end{aligned}
\]
\end{proof}

The two error terms have different sources. In the ideal intervention model, \(\rho_{\mathrm{err}}\) may be set to \(\|\sum_a\pi_a\Omega_a\|_{\mathrm{op}}\); \cref{prop:intervention-error} also bounds target leakage and unintended changes. Under the fixed iid-law assumptions of \cref{thm:voirdire-sample}, its concentration bound supplies \(\epsilon_M\). The deterministic statement applies to fixed family counts as well, but its probability guarantee requires a concentration bound for that sampling design.

The target \(L_C\) measures nuisance variation excited by the chosen interventions. It need not equal the nuisance covariance of the original items. Accurate estimation of its normalized matrix does not imply strong attenuation of every nuisance direction: \cref{thm:attenuation} separately controls that effect through the ratio of the weakest to the strongest nuisance eigenvalue.

\subsubsection{Subspace recovery as an additional consequence}
\label{app:paired-subspace}

Subspace recovery requires a spectral gap in addition to matrix accuracy. We express it using the orthogonal projector \(P_{\mathcal S}\) onto \(\mathcal S\). For full-column-rank \(U_C\), \(P_{U_C}=U_C(U_C^\top U_C)^{-1}U_C^\top\), reducing to \(U_CU_C^\top\) for orthonormal columns. For equidimensional subspaces,
\begin{equation}\label{eq:projector-sin-theta}
    \|P_{\mathcal A}-P_{\mathcal B}\|_{\mathrm{op}}
    =\|\sin\Theta(\mathcal A,\mathcal B)\|_{\mathrm{op}}.
\end{equation}
These projectors describe subspace diagnostics; the algorithm uses the normalized moment \(\widehat T\).

\begin{corollary}[Nuisance-subspace perturbation]\label{cor:paired-subspace}
Under the ideal intervention setup of \cref{prop:paired-difference}, assume \(U_C\) has full column rank \(1\le d_C\le p\). If \(\Omega_a=0\) and \(A_a\) has rank \(d_C\), then \(\operatorname{range}(M_a)=\operatorname{span}(U_C)\). With paired judge noise, define
\[
\begin{aligned}
    L_a &:= U_C A_a U_C^\top,\\
    \lambda_{\mathrm{int}} &:= \lambda_{d_C}(L_a),\\
    \rho_{\mathrm{pair}} &:= \|\Omega_a\|_{\mathrm{op}},
\end{aligned}
\]
where \(\lambda_j(M)\) denotes the \(j\)-th largest eigenvalue of a symmetric matrix \(M\). Thus \(\lambda_{\mathrm{int}}\) is the weakest nuisance-signal eigenvalue in this model. If \(\lambda_{\mathrm{int}}>2\rho_{\mathrm{pair}}\), then the top-\(d_C\) eigenspace \(\mathcal C_a\) of \(M_a\) satisfies
\[
    \|P_{\mathcal C_a}-P_{U_C}\|_{\mathrm{op}}
    \le
    \frac{2\rho_{\mathrm{pair}}}{\lambda_{\mathrm{int}}}.
\]
Since \(U_C\mu_a\mu_a^\top U_C^\top\) is positive semidefinite, \(\lambda_{\mathrm{int}}\) is at least the value obtained from \(\Cov(\delta_i^{(a)})\) alone: a systematic shift cannot decrease the weakest nuisance eigenvalue and may supply a direction that item-to-item variation does not.
\end{corollary}

\begin{proof}
By \cref{eq:paired-covariance-minimal}, if \(\Omega_a=0\), then
\[
    M_a = U_CA_aU_C^\top .
\]
Because \(U_C\) has full column rank and \(A_a\) has rank \(d_C\), this matrix has range \(\operatorname{span}(U_C)\).

It remains to handle paired judge noise. Write
\[
\begin{aligned}
    G_a &:= M_a,\\
    G_a &= L_a+\Omega_a,\\
    L_a &= U_CA_aU_C^\top,\\
    \Omega_a &= \Cov(\xi_i^{(a)}).
\end{aligned}
\]
If \(d_C=p\), both subspaces are the full judge-output space and their projectors agree. Suppose now that \(d_C<p\). The matrix \(L_a\) has rank \(d_C\), range \(\operatorname{span}(U_C)\), and eigengap \(\lambda_{d_C}(L_a)\) between its \(d_C\)-th and \((d_C+1)\)-st eigenvalues. By Weyl's inequality and the Davis--Kahan \(\sin\Theta\) theorem \citep{bhatia1997matrix,davis1970rotation,yu2015useful}, if \(\lambda_{d_C}(L_a)>2\|\Omega_a\|_{\mathrm{op}}\), then the top-\(d_C\) eigenspace \(\mathcal C_a\) of \(G_a\) satisfies
\[
\begin{aligned}
    &\|P_{\mathcal C_a}-P_{U_C}\|_{\mathrm{op}} \\
    &\quad\le
    \frac{\|\Omega_a\|_{\mathrm{op}}}
         {\lambda_{d_C}(L_a)-\|\Omega_a\|_{\mathrm{op}}} \\
    &\quad\le
    \frac{2\|\Omega_a\|_{\mathrm{op}}}{\lambda_{d_C}(L_a)} ,
\end{aligned}
\]
using \cref{eq:projector-sin-theta} to read the Davis--Kahan conclusion as a projector distance. Substituting \(\lambda_{\mathrm{int}}=\lambda_{d_C}(L_a)\) and \(\rho_{\mathrm{pair}}=\|\Omega_a\|_{\mathrm{op}}\) gives the stated bound. Finally, \(L_a=U_C\Cov(\delta_i^{(a)})U_C^\top+U_C\mu_a\mu_a^\top U_C^\top\) is a sum of positive semidefinite matrices, so by Weyl's monotonicity \(\lambda_{d_C}(L_a)\ge\lambda_{d_C}(U_C\Cov(\delta_i^{(a)})U_C^\top)\), the value obtained from the centered covariance.
\end{proof}

The proof formalizes the paired-comparison intuition. If the rewrite preserves the target, then subtracting the original and rewritten judge outputs removes the target part. The remaining consistent score changes show which judge-output directions react to the rewrite; these are the directions \voirdire{} later attenuates. Paired judge noise can blur this signal, but the corollary shows recovery works when the rewrite-induced changes are larger than that noise.

\subsection{Effect of imperfect interventions}

The rewrites are intended to change only the specified nuisance attribute while preserving the evaluation target. In practice, a rewrite that passes the checks may deviate from this ideal: it may accidentally change target-relevant content, or it may introduce additional changes beyond the intended nuisance attribute. We treat these deviations as perturbations to the paired-difference second moment.

We decompose the paired difference as
\begin{equation}\label{eq:intervention-error-paired-difference}
    z_i^{(a)}
    =
    U_C\bar\delta_i^{(a)}
    +
    U_Q r_i^{(a)}
    +
    U_E e_i^{(a)}
    +
    \xi_i^{(a)} .
\end{equation}
Here \(\bar\delta_i^{(a)}\) is the intended nuisance movement. The term \(r_i^{(a)}\) captures accidental changes to the target factor, and \(U_E e_i^{(a)}\) captures rewrite-induced changes outside the intended nuisance subspace. The residual term \(\xi_i^{(a)}\) is paired judge noise. Thus imperfect rewrites add error terms to the paired-difference second moment, which can interfere with recovery of the nuisance-sensitive subspace.

\begin{proposition}[Perturbation from imperfect interventions]\label{prop:intervention-error}
Fix a family \(a\) and let
\[
\begin{aligned}
    L_C&=U_C\,\mathbb E\bigl[\bar\delta_i^{(a)}\bar\delta_i^{(a)\top}\bigr]U_C^\top,\\
    \mathbb E\bigl[z_i^{(a)}z_i^{(a)\top}\bigr]&=L_C+E_{\mathrm{err}} .
\end{aligned}
\]
If \(L_C\) has rank \(d_C\), where \(1\le d_C\le p\), \(\lambda_\star=\lambda_{d_C}(L_C)\), and \(\|E_{\mathrm{err}}\|_{\mathrm{op}}\le\rho_{\mathrm{err}}<\lambda_\star/2\), then the top-\(d_C\) eigenspace \(\mathcal C\) of \(\mathbb E[z_i^{(a)}z_i^{(a)\top}]\) satisfies
\[
    \|P_{\mathcal C}-P_{U_C}\|_{\mathrm{op}}
    \le
    \frac{2\rho_{\mathrm{err}}}{\lambda_\star} .
\]
Moreover, let \(D,R,B\ge0\) and \(\pi_Q,\pi_E\in[0,1]\). Suppose \(U_Q,U_C,U_E\) have orthonormal columns, \(\xi_i^{(a)}\) is centered and uncorrelated with \(\bar\delta_i^{(a)},r_i^{(a)},e_i^{(a)}\), and
\[
    \|\Cov(\xi_i^{(a)})\|_{\mathrm{op}}\le\rho_{\mathrm{pair}},
    \qquad
    \mathbb E\|\bar\delta_i^{(a)}\|_2^2\le D^2,
\]
\[
    \mathbb E\|r_i^{(a)}\|_2^2\le \pi_QR^2,
    \qquad
    \mathbb E\|e_i^{(a)}\|_2^2\le \pi_EB^2 .
\]
Then the perturbation itself satisfies
\[
\begin{aligned}
    \|E_{\mathrm{err}}\|_{\mathrm{op}}
    \le{}&
    \rho_{\mathrm{pair}}
    +\pi_QR^2
    +\pi_EB^2 \\
    &+2D(\sqrt{\pi_Q}R+\sqrt{\pi_E}B)
    +2\sqrt{\pi_Q\pi_E}RB .
\end{aligned}
\]
\end{proposition}

\begin{proof}
The first claim follows from a standard perturbation argument. By definition,
\[
    \mathbb E\bigl[z_i^{(a)}z_i^{(a)\top}\bigr]=L_C+E_{\mathrm{err}}.
\]
The matrix \(L_C\) is the second moment contributed by the intended nuisance movement. If \(d_C=p\), both projectors equal \(I\), so the recovery bound is immediate. For \(d_C<p\), it has rank \(d_C\) and eigengap \(\lambda_\star\) between its \(d_C\)-th and \((d_C+1)\)-st eigenvalues. Since \(\|E_{\mathrm{err}}\|_{\mathrm{op}}<\lambda_\star/2\), the Davis--Kahan \(\sin\Theta\) theorem \citep{davis1970rotation,yu2015useful} gives
\[
    \|P_{\mathcal C}-P_{U_C}\|_{\mathrm{op}}
    \le
    \frac{\rho_{\mathrm{err}}}{\lambda_\star-\rho_{\mathrm{err}}}
    \le
    \frac{2\rho_{\mathrm{err}}}{\lambda_\star} .
\]

For the second claim, expand the second moment of \cref{eq:intervention-error-paired-difference}. The error term contains paired judge noise, target-changing terms, unintended rewrite terms, and their cross moments with the intended nuisance movement; the cross moments with the centered paired noise vanish as in \cref{prop:paired-difference}. Since \(U_Q,U_C,U_E\) have orthonormal columns, multiplication by these loading matrices does not increase operator norm. Therefore,
\[
\begin{aligned}
    \bigl\|\mathbb E[r_i^{(a)}r_i^{(a)\top}]\bigr\|_{\mathrm{op}}
    &\le
    \mathbb E\|r_i^{(a)}\|_2^2
    \le
    \pi_QR^2,\\
    \bigl\|\mathbb E[e_i^{(a)}e_i^{(a)\top}]\bigr\|_{\mathrm{op}}
    &\le
    \mathbb E\|e_i^{(a)}\|_2^2
    \le
    \pi_EB^2 .
\end{aligned}
\]
By Cauchy--Schwarz,
\[
\begin{aligned}
    \bigl\|\mathbb E[\bar\delta_i^{(a)}r_i^{(a)\top}]\bigr\|_{\mathrm{op}}
    &\le
    D\sqrt{\pi_Q}R,\\
    \bigl\|\mathbb E[\bar\delta_i^{(a)}e_i^{(a)\top}]\bigr\|_{\mathrm{op}}
    &\le
    D\sqrt{\pi_E}B,
\end{aligned}
\]
and
\[
    \bigl\|\mathbb E[r_i^{(a)}e_i^{(a)\top}]\bigr\|_{\mathrm{op}}
    \le
    \sqrt{\pi_Q\pi_E}RB .
\]
These uncentered cross moments already contain the products of the means, so no separate centering term arises. Each cross moment appears together with its transpose, giving the factors of two. Adding the paired-noise covariance yields the displayed bound on \(\|E_{\mathrm{err}}\|_{\mathrm{op}}\). Its right-hand side may be used as \(\rho_{\mathrm{err}}\); the subspace conclusion additionally requires this value to be less than \(\lambda_\star/2\).
\end{proof}

This proposition clarifies the role of the preservation checks. The checks do not need to eliminate all errors, but they should keep target-changing and unintended rewrite effects small relative to the intended nuisance signal. Recovery is reliable when the intended nuisance eigengap \(\lambda_\star\) dominates the combined perturbation level \(\rho_{\mathrm{err}}\).

\subsection{Finite-sample recovery and effective rank}
The preceding results use population moments. In the algorithm, however, the intervention second moment is estimated from finitely many calibration pairs. The next result separates two sources of error: the perturbation from imperfect interventions, measured by \(\rho_{\mathrm{err}}\), and the moment-estimation error, measured by \(\epsilon_M\). The latter depends on both the pooled calibration-pair count \(M\) and the effective rank \(r_{\mathrm{eff}}(G)\) of the paired-difference second moment.

\begin{theorem}[Finite-sample nuisance recovery]\label{thm:voirdire-sample}
Let \(\bar G=\frac1M\sum_{i=1}^M z_iz_i^\top\) be the pooled second moment of the \(M\) calibration paired differences (\cref{eq:family-shift-covariance}), with \(M\ge1\). Let \(G:=\mathbb E[zz^\top]\) denote the population pooled moment. Suppose
\[
\begin{aligned}
    G &= L_C+E_{\mathrm{err}},\\
    \|E_{\mathrm{err}}\|_{\mathrm{op}} &\le \rho_{\mathrm{err}},\\
    \|\bar G-G\|_{\mathrm{op}} &\le \epsilon_M .
\end{aligned}
\]
If \(L_C\succeq0\) has rank \(d_C\), where \(1\le d_C\le p\), \(\operatorname{range}(L_C)=\operatorname{span}(U_C)\), and
\[
    \lambda_\star
    :=
    \lambda_{d_C}(L_C)
    >
    2(\rho_{\mathrm{err}}+\epsilon_M),
\]
then any orthonormal basis matrix \(\widehat U_C\) for the top-\(d_C\) eigenspace of \(\bar G\) satisfies
\[
    \|P_{\widehat U_C}-P_{U_C}\|_{\mathrm{op}}
    \le
    \frac{2(\rho_{\mathrm{err}}+\epsilon_M)}{\lambda_\star} .
\]

For the probability bound, additionally suppose the \(z_i\) are iid from a fixed pooled distribution, \(G\ne0\), and, for a fixed \(K\ge1\),
\[
    \|u^\top z\|_{\psi_2}\le K\sqrt{u^\top G u}
    \qquad\text{for every }u\in\mathbb R^p.
\]
Here \(\|Y\|_{\psi_2}:=\inf\{t>0:\mathbb E\exp(Y^2/t^2)\le2\}\). Then, for \(\delta\in(0,1)\), with probability at least \(1-\delta\),
\[
\begin{aligned}
    \|\bar G-G\|_{\mathrm{op}}
    \le{}&
    C_K\|G\|_{\mathrm{op}}
    \Bigg(
        \sqrt{\frac{r_{\mathrm{eff}}(G)+\log(1/\delta)}{M}} \\
        &\qquad\qquad+
        \frac{r_{\mathrm{eff}}(G)+\log(1/\delta)}{M}
    \Bigg),
\end{aligned}
\]
where \(r_{\mathrm{eff}}(G)=\operatorname{tr}(G)/\|G\|_{\mathrm{op}}\) and \(C_K\) depends only on \(K\).
\end{theorem}

\begin{proof}
Let
\[
    G=\mathbb E[zz^\top]=L_C+E_{\mathrm{err}} .
\]
Then
\[
    \bar G-L_C
    =
    E_{\mathrm{err}}+(\bar G-G).
\]
By the triangle inequality,
\[
    \|\bar G-L_C\|_{\mathrm{op}}
    \le
    \rho_{\mathrm{err}}+\epsilon_M .
\]
If \(d_C=p\), both projectors equal \(I\). For \(d_C<p\), since \(L_C\) has rank \(d_C\) and \(\lambda_\star>2(\rho_{\mathrm{err}}+\epsilon_M)\), Davis--Kahan gives
\[
\begin{aligned}
    &\|P_{\widehat U_C}-P_{U_C}\|_{\mathrm{op}}\\
    &\quad\le
    \frac{\rho_{\mathrm{err}}+\epsilon_M}
         {\lambda_\star-\rho_{\mathrm{err}}-\epsilon_M}\\
    &\quad\le
    \frac{2(\rho_{\mathrm{err}}+\epsilon_M)}{\lambda_\star}.
\end{aligned}
\]

For the probability bound, let \(s_i\) be independent Rademacher signs, independent of the observations, and set \(y_i=s_i z_i\). Then \(\mathbb E[y_i]=0\), \(y_i y_i^\top=z_i z_i^\top\), and \(\|u^\top y_i\|_{\psi_2}=\|u^\top z_i\|_{\psi_2}\). Thus the sample second moment is unchanged, while the \(y_i\) meet the centered, uniformly sub-gaussian hypotheses of the effective-rank covariance bound \citep[Theorem~9]{koltchinskii2017concentration}. Applying that bound on \(\operatorname{range}(G)\) gives the result, with constants depending only on \(K\).
\end{proof}

The deterministic recovery bound does not require independent samples once $\epsilon_M$ is bounded. The probability guarantee assumes iid sampling from a fixed pooled distribution; fixed per-family counts or pairs sharing an original item require a corresponding concentration argument. If $G=0$, the estimation error vanishes, but the positive nuisance-eigengap condition cannot hold.

Increasing $M$ reduces the estimation error $\epsilon_M$ at a rate governed by the effective rank $r_{\mathrm{eff}}(G)$: fewer pairs are needed when paired-difference variation is concentrated in a few directions. However, additional pairs cannot eliminate the systematic perturbation $\rho_{\mathrm{err}}$. Accurate recovery therefore requires the nuisance signal $\lambda_\star$ to dominate both sources of error.

\subsection{Effect of attenuation}

The calibration step applies the operator \(I-\alpha\widehat T\) to each judge-output vector, where \(\widehat T=\bar G/\lambda_{\max}(\bar G)\) is the normalized pooled shift moment (\cref{eq:spectral-calibration}). In the ideal orthogonal model below, this step has a simple tradeoff: increasing \(\alpha\) attenuates more nuisance variation, while error in estimating \(T\) can also perturb the target component.

\begin{theorem}[Effect of attenuation]\label{thm:attenuation}
Assume the noiseless orthogonal model
\[
    x_i^{(0)}=U_Q q_i+U_C c_i,
    \qquad
    x_i^{(a)}=x_i^{(0)}+U_C\delta_i^{(a)},
\]
where \(U_Q\) and \(U_C\) have orthonormal columns and \(U_Q^\top U_C=0\). Let \(A=\mathbb E[\delta_i^{(a)}\delta_i^{(a)\top}]\) have rank \(d_C\) with eigenvalues \(\lambda_1\ge\dots\ge\lambda_{d_C}>0\), let \(L_C=U_CAU_C^\top\) and \(T=L_C/\lambda_1\), and write \(\kappa:=\lambda_{d_C}/\lambda_1\in(0,1]\) for the spectral ratio of the nuisance movement. Suppose
\[
    \|\widehat T-T\|_{\mathrm{op}}\le \eta .
\]
For \(\alpha\in[0,1]\), define
\[
\begin{aligned}
    \widetilde x_i(\alpha)
    &=(I-\alpha\widehat T)x_i^{(0)},\\
    \widetilde x_i^{(a)}(\alpha)
    &=(I-\alpha\widehat T)x_i^{(a)} .
\end{aligned}
\]
Then
\[
\begin{aligned}
    &\|\widetilde x_i(\alpha)-U_Q q_i\|_2\\
    &\quad\le
    (1-\alpha\kappa)\|c_i\|_2
    +
    \alpha\eta(\|q_i\|_2+\|c_i\|_2),
\end{aligned}
\]
and
\[
\begin{aligned}
    &\|\widetilde x_i^{(a)}(\alpha)-\widetilde x_i(\alpha)\|_2\\
    &\quad\le
    \bigl((1-\alpha\kappa)+\alpha\eta\bigr)
    \|\delta_i^{(a)}\|_2 .
\end{aligned}
\]
If the calibration differences carry centered judge noise uncorrelated with the nuisance movement, with covariance \(\Omega\), the population moment is \(G=L_C+\Omega\). For \(\widehat T=G/\lambda_{\max}(G)\), \(\|\widehat T-T\|_{\mathrm{op}}\le 2\|\Omega\|_{\mathrm{op}}/\lambda_1\), so this value may be used for \(\eta\).
\end{theorem}

\begin{proof}
Because \(U_Q^\top U_C=0\), the operator satisfies
\[
    TU_Q=0,
    \qquad
    TU_C=U_CA/\lambda_1 .
\]
The matrix \(D_\alpha:=I-\alpha A/\lambda_1\) is symmetric with eigenvalues \(1-\alpha\lambda_j/\lambda_1\in[1-\alpha,\,1-\alpha\kappa]\), so \(\|D_\alpha v\|_2\le(1-\alpha\kappa)\|v\|_2\) for every \(v\). For the first inequality,
\[
\begin{aligned}
    &\widetilde x_i(\alpha)-U_Qq_i\\
    &\quad=
    (I-\alpha\widehat T)(U_Qq_i+U_Cc_i)-U_Qq_i\\
    &\quad=
    -\alpha(\widehat T-T)U_Qq_i\\
    &\qquad
    +U_CD_\alpha c_i\\
    &\qquad
    +\alpha(T-\widehat T)U_Cc_i .
\end{aligned}
\]
Taking norms, using \(\|\widehat T-T\|_{\mathrm{op}}\le\eta\), and using the orthonormality of \(U_Q\) and \(U_C\) gives
\[
\begin{aligned}
    &\|\widetilde x_i(\alpha)-U_Qq_i\|_2\\
    &\le
    (1-\alpha\kappa)\|c_i\|_2
    +
    \alpha\eta(\|q_i\|_2+\|c_i\|_2).
\end{aligned}
\]

For a calibration pair under this target-invariance assumption,
\[
    x_i^{(a)}-x_i^{(0)}=U_C\delta_i^{(a)} .
\]
Therefore
\[
\begin{aligned}
    &\widetilde x_i^{(a)}(\alpha)-\widetilde x_i(\alpha)\\
    &\quad=
    (I-\alpha\widehat T)U_C\delta_i^{(a)}\\
    &\quad=
    U_CD_\alpha\delta_i^{(a)}
    +
    \alpha(T-\widehat T)U_C\delta_i^{(a)} .
\end{aligned}
\]
The same norm bound gives
\[
\begin{aligned}
    \|\widetilde x_i^{(a)}(\alpha)-\widetilde x_i(\alpha)\|_2
    \le
    \bigl((1-\alpha\kappa)+\alpha\eta\bigr)
    \|\delta_i^{(a)}\|_2 .
\end{aligned}
\]

For the noise claim, \(L_C\) and \(\Omega\) are positive semidefinite, so \(\lambda_{\max}(L_C+\Omega)\ge\lambda_{\max}(L_C)=\lambda_1>0\). Apply \cref{eq:normalized-nuisance-error} with \(\bar G=G=L_C+\Omega\), \(\rho_{\mathrm{err}}=\|\Omega\|_{\mathrm{op}}\), and \(\epsilon_M=0\). The same result supplies \(\eta\) with an additional sampling term when the moment is empirical.
\end{proof}

The theorem captures the attenuation tradeoff. If \(T\) is estimated accurately, increasing \(\alpha\) reduces both the remaining nuisance component and the original--rewrite score change, and each nuisance direction is scaled by a factor in \([1-\alpha,1-\alpha\kappa]\) according to how strongly the rewrites move it. If \(T\) is inaccurate, larger \(\alpha\) can also remove target-relevant variation, which is why \(\alpha\) is selected on validation rather than fixed at one. When \(A\) is a multiple of the identity, \(T=P_{U_C}\) is the projector onto the nuisance subspace, \(\kappa=1\), and the operator becomes projector-based attenuation; it removes the nuisance component completely when \(\alpha=1\) and \(\widehat T=T\).

\subsection{Perturbation of the generalized eigenvector}
\label{app:gevd-perturbation}

We study the leading generalized eigenvector of the calibrated-output covariance and intervention penalty. Write \(\Sigma=\Sigma_\alpha\succeq0\) and \(B=\bar G+\rho I\succ0\). The weight vector solves
\begin{equation}\label{eq:gevd}
    \Sigma w=\mu B w.
\end{equation}
The result below compares unit-normalized weights up to relative sign in the norm \(\|v\|_B=(v^\top Bv)^{1/2}\). The implementation's orientation and normalization require the additional local conditions stated afterward.

\begin{theorem}[Generalized eigenvector perturbation]
\label{thm:gevd-perturbation}
Let \(\Sigma,\widehat\Sigma\succeq0\) and \(B,\widehat B\succ0\) be \(p\times p\) matrices, with \(p\ge2\) and \(B\succeq\rho I\) for \(\rho>0\). Suppose
\[
    \|\widehat\Sigma-\Sigma\|_{\mathrm{op}}\le\epsilon_\Sigma,
    \qquad
    \|\widehat B-B\|_{\mathrm{op}}\le\epsilon_B<\rho.
\]
Let \(\mu_1\) and \(\Delta>0\) be the largest generalized eigenvalue and the gap to the second largest eigenvalue of \((\Sigma,B)\). Define
\[
    \tau:=\epsilon_B/\rho,
    \qquad
    \zeta:=\frac{\epsilon_\Sigma+\mu_1\epsilon_B}{\rho-\epsilon_B}.
\]
If \(\Delta>2\zeta\), then leading generalized eigenvectors \(w,\hat w\) normalized by \(\|w\|_B=\|\hat w\|_{\widehat B}=1\) satisfy
\begin{equation}\label{eq:gevd-bound}
\begin{aligned}
    \min_{s\in\{-1,1\}}\|s\hat w-w\|_B
    \le{}&\frac{2\sqrt2\,\zeta}{\Delta\sqrt{1-\tau}}\\
    &+\frac{1}{\sqrt{1-\tau}}-1.
\end{aligned}
\end{equation}
\end{theorem}

\begin{proof}
Use the same reference coordinates for both pencils. Set
\[
\begin{aligned}
    M&=B^{-1/2}\Sigma B^{-1/2},\\
    E&=B^{-1/2}(\widehat B-B)B^{-1/2},\\
    F&=B^{-1/2}(\widehat\Sigma-\Sigma)B^{-1/2},\\
    S&=(I+E)^{-1/2}.
\end{aligned}
\]
Then \(\|E\|_{\mathrm{op}}\le\tau<1\), \(\|F\|_{\mathrm{op}}\le\epsilon_\Sigma/\rho\), and \(\|M\|_{\mathrm{op}}=\mu_1\). The eigenvalues of \(S\) lie between \((1+\tau)^{-1/2}\) and \(a:=(1-\tau)^{-1/2}\), so \(\|S\|_{\mathrm{op}}\le a\) and \(\|S-I\|_{\mathrm{op}}\le a-1\). The second inequality uses \(1-(1+\tau)^{-1/2}\le(1-\tau)^{-1/2}-1\).

The transformed estimated pencil is the symmetric matrix \(N=S(M+F)S\). Its eigenvalues are those of \((\widehat\Sigma,\widehat B)\). Expanding the difference gives
\[
    N-M=SFS+(S-I)MS+M(S-I),
\]
and hence
\[
\begin{aligned}
    \|N-M\|_{\mathrm{op}}
    &\le a^2\epsilon_\Sigma/\rho+\mu_1(a-1)(a+1)\\
    &=\zeta.
\end{aligned}
\]
Let \(y=B^{1/2}w\) and \(z=(I+E)^{1/2}B^{1/2}\hat w\). These are unit leading eigenvectors of \(M\) and \(N\), respectively. Weyl's inequality and \(\Delta>2\zeta\) ensure that the estimated leading eigenvalue is simple. The single-vector Davis--Kahan bound gives a relative sign \(s\) such that \(\|sz-y\|_2\le2\sqrt2\,\zeta/\Delta\) \citep{yu2015useful}.

Mapping back now retains the change in the metric:
\[
    B^{1/2}(s\hat w-w)=S(sz-y)+(S-I)y.
\]
Taking norms and using \(\|y\|_2=1\) gives \(\|s\hat w-w\|_B\le a\,2\sqrt2\,\zeta/\Delta+(a-1)\), which is \cref{eq:gevd-bound}.
\end{proof}

\paragraph{Ridge perturbation.}
The bound controls the entire denominator matrix. Let \(\bar G'\) denote a second empirical moment. With a fixed ridge, \(\epsilon_B\) may be taken as \(\|\bar G'-\bar G\|_{\mathrm{op}}\). For the implemented rule \(\rho=\gamma\operatorname{tr}(\bar G)/p\) with fixed \(\gamma>0\), recomputing the ridge gives
\[
    \|\widehat B-B\|_{\mathrm{op}}
    \le(1+\gamma)\|\bar G'-\bar G\|_{\mathrm{op}}.
\]
This follows from \(|\operatorname{tr}(H)|\le p\|H\|_{\mathrm{op}}\) for symmetric \(H\). The result compares a fixed estimator configuration and family set; it does not bound changes caused by discrete validation selection.

\paragraph{Transfer to reported scores.}
For a fixed calibrated-output matrix \(\widetilde X\) and relatively aligned, unit-normalized weights, the bound gives
\[
    \|\widetilde X(s\hat w-w)\|_2
    \le\|\widetilde X\|_{\mathrm{op}}\rho^{-1/2}\|s\hat w-w\|_B.
\]
The implementation instead orients the weights to a positive coordinate sum and rescales them to sum to one (\cref{app:current-protocol}). This map is not continuous where the coordinate sum is zero.

A local transfer is available when relatively aligned weights \(u,v\) have the same orientation. Write the normalization as \(\mathcal N(v)=v/d(v)\) with \(d(v)=\mathbf 1^\top v\). If \(d(u),d(v)\ge b>0\) and \(\|u\|_2\le R\), then
\[
\begin{aligned}
    \|\mathcal N(v)-\mathcal N(u)\|_2
    &\le\left(\frac1b+\frac{R\sqrt p}{b^2}\right)\|v-u\|_2.
\end{aligned}
\]
Indeed, \(|d(v)-d(u)|\le\sqrt p\|v-u\|_2\), and
\[
    \frac{v}{d(v)}-\frac{u}{d(u)}
    =\frac{v-u}{d(v)}
      +u\frac{d(u)-d(v)}{d(u)d(v)}.
\]
A strict margin from the coordinate-sum sign boundary ensures the local orientation condition and the lower bound \(b\). Put \(L_{b,R}=1/b+R\sqrt p/b^2\). If recalibration also changes \(\widetilde X\) to \(\widetilde X'\), the full score change satisfies
\[
\begin{aligned}
    &\|\widetilde X'\mathcal N(v)-\widetilde X\mathcal N(u)\|_2\\
    &\quad\le\|\widetilde X\|_{\mathrm{op}}L_{b,R}\|v-u\|_2\\
    &\qquad+\|\widetilde X'-\widetilde X\|_{\mathrm{op}}
        \|\mathcal N(v)\|_2.
\end{aligned}
\]
This follows by adding and subtracting \(\widetilde X\mathcal N(v)\). For fixed original outputs \(X\) and \(\alpha\in[0,1]\), the two calibration operators give \(\widetilde X=X(I-\alpha\widehat T)\) and \(\widetilde X'=X(I-\alpha\widehat T')\), so
\[
    \|\widetilde X'-\widetilde X\|_{\mathrm{op}}
    \le\alpha\|X\|_{\mathrm{op}}
        \|\widehat T'-\widehat T\|_{\mathrm{op}}.
\]
For the same \(n\ge1\) evaluation items and targets, the absolute MAE change is at most \(n^{-1/2}\) times the full score-vector bound. Thus both changing weights and changing calibration contribute. Without the orientation margin, the unit-normalized eigenvector bound does not imply stability of the reported scores.

\paragraph{Target and nuisance blocks.}
Consider the population orthogonal model of \cref{thm:attenuation} with \(\bar G=L_C=U_CAU_C^\top\), so \(\widehat T=T\). This assumes the unnormalized moment is exact; equality of the normalized operators alone does not fix its scale. Both matrices of the pencil have a block form. Writing \(\Sigma_{QQ}=\Cov(q_i)\), \(\Sigma_{QC}=\Cov(q_i,c_i)\), \(\Sigma_{CC}=\Cov(c_i)\), and \(D_\alpha=I-\alpha A/\lambda_1\),
\[
\begin{aligned}
    \Sigma_\alpha
    ={}&U_Q\Sigma_{QQ}U_Q^\top
    +U_Q\Sigma_{QC}D_\alpha U_C^\top\\
    &+U_CD_\alpha\Sigma_{QC}^\top U_Q^\top
    +U_CD_\alpha\Sigma_{CC}D_\alpha U_C^\top ,
\end{aligned}
\]
so attenuation leaves the target block unchanged and shrinks the nuisance block by \(D_\alpha\) on both sides, while \(B=\bar G+\rho I\) equals \(\rho I\) on \(\operatorname{span}(U_Q)\) and \(U_C(A+\rho I)U_C^\top\) on \(\operatorname{span}(U_C)\). The generalized eigenvector therefore weights target variance against \(\rho\) alone and nuisance variance against \(A+\rho I\): a direction that the rewrites move strongly is penalized in the denominator in addition to being attenuated in the numerator.

\subsection{Pooled second moment across families}
\label{app:shared-intervention-covariance}

Proposition~\ref{prop:paired-difference} already controls the pooled moment used by \voirdire{}. The next corollary gives its additional subspace-recovery consequence when a fixed set of families perturbs a common nuisance subspace, possibly with different strengths and systematic shifts.

\begin{corollary}[Pooled second moment]
\label{cor:shared-intervention-covariance}
Fix a nonempty family set \(\mathcal A\) and positive calibration counts \(m_a\), with \(M=\sum_{a\in\mathcal A}m_a\) and \(\pi_a=m_a/M\). Suppose each family satisfies
\[
    \mathbb E\bigl[z_i^{(a)}z_i^{(a)\top}\bigr]
    =
    U_CA_aU_C^\top+\Omega_a,
    \qquad a\in\mathcal A .
\]
Define the population pooled moment and its signal/noise decomposition by
\[
\begin{aligned}
    G_{\mathrm{pool}}
    &=
    \sum_{a\in\mathcal A}\pi_a\,\mathbb E\bigl[z_i^{(a)}z_i^{(a)\top}\bigr],\\
    L_{\mathrm{pool}}
    &=
    U_C
    \left(
        \sum_{a\in\mathcal A}\pi_aA_a
    \right)
    U_C^\top,\\
    \Omega_{\mathrm{pool}}
    &=
    \sum_{a\in\mathcal A}\pi_a\Omega_a .
\end{aligned}
\]
If \(L_{\mathrm{pool}}\) has rank \(d_C\), where \(1\le d_C\le p\),
\[
\begin{aligned}
    \lambda_{\mathrm{pool}}
    &:=
    \lambda_{d_C}(L_{\mathrm{pool}}),\\
    \rho_{\mathrm{pool}}
    &:=
    \|\Omega_{\mathrm{pool}}\|_{\mathrm{op}}
    <
    \lambda_{\mathrm{pool}}/2,
\end{aligned}
\]
then the top-\(d_C\) eigenspace \(\mathcal C_{\mathrm{pool}}\) of \(G_{\mathrm{pool}}\) satisfies
\[
    \|P_{\mathcal C_{\mathrm{pool}}}-P_{U_C}\|_{\mathrm{op}}
    \le
    \frac{2\rho_{\mathrm{pool}}}{\lambda_{\mathrm{pool}}}.
\]
\end{corollary}

\begin{proof}
The pooled estimator \(\bar G=\frac1M\sum_{i=1}^Mz_iz_i^\top\) gives every calibration pair equal weight, so its expectation is
\[
    G_{\mathrm{pool}}
    =
    L_{\mathrm{pool}}+\Omega_{\mathrm{pool}} .
\]
The signal matrix \(L_{\mathrm{pool}}\) has range \(\operatorname{span}(U_C)\). If \(d_C=p\), both projectors equal \(I\). For \(d_C<p\), the signal has eigengap \(\lambda_{\mathrm{pool}}\) between its \(d_C\)-th and \((d_C+1)\)-st eigenvalues. Applying the Davis--Kahan argument from \cref{cor:paired-subspace} yields
\[
\begin{aligned}
    \|P_{\mathcal C_{\mathrm{pool}}}-P_{U_C}\|_{\mathrm{op}} \le
    \frac{\rho_{\mathrm{pool}}}
         {\lambda_{\mathrm{pool}}-\rho_{\mathrm{pool}}} \le
    \frac{2\rho_{\mathrm{pool}}}{\lambda_{\mathrm{pool}}}.
\end{aligned}
\]
\end{proof}

This is the population analogue of the pooled \voirdire{} estimator: the algorithm stacks the paired shifts from all selected families, estimates one second moment, and uses it to form \(\widehat T\) and the penalty in \cref{eq:aggregation-objective}. The family means \(\mu_a\) enter only through \(A_a=\Cov(\delta_i^{(a)})+\mu_a\mu_a^\top\), so a family whose shift is mostly systematic contributes a rank-one term inside \(\operatorname{span}(U_C)\). Pooling strengthens nuisance-subspace recovery when the added families increase the nuisance signal relative to perturbation. The benefit depends on the strength and diversity of the intervention-induced shifts.
\section{Synthetic Experiments}
\label{app:synthetic-theory-stress}

We use synthetic data with known target and nuisance components to examine how intervention errors, attenuation strength, and systematic score shifts affect calibration (\cref{app:synthetic-error-floor,app:synthetic-attenuation,app:synthetic-mean-shift}). We then examine how nuisance-subspace recovery relates to aggregation performance (\cref{app:synthetic-clean-recovery}).

\paragraph{Common setup.}
Unless otherwise stated, we use \(p=24\) judges and two-dimensional target and nuisance factors (\(d_Q=d_C=2\)). The combined loading matrix \([U_Q,U_C]\) has orthonormal columns. For experiments involving random sampling, we report means over 20 random seeds, with shaded bands indicating one standard error.

\subsection{Sensitivity to Imperfect Interventions}
\label{app:synthetic-error-floor}
We examine how unintended changes to target quality affect recovery of nuisance-sensitive directions.

\paragraph{Setup.}
Following \cref{prop:intervention-error}, we generate paired shifts as
\[
    z=U_C\delta+U_Qr+\xi,\qquad
    \delta=\begin{pmatrix}0.6\\-0.3\end{pmatrix}
    +\operatorname{diag}(2,1)g,\qquad
    r=b_Qa_Q\left(g+\begin{pmatrix}0.2\\-0.1\end{pmatrix}\right),
\]
where \(g\sim N(0,I_2)\), \(b_Q\sim\operatorname{Bernoulli}(\pi_Q)\), and \(\xi\sim N(0,0.02I_p)\) are independent. Here, \(\pi_Q\) controls how often interventions change the target, and \(a_Q\) controls the magnitude of those changes. The shared \(g\) makes target and nuisance changes correlated. We vary \(\pi_Q\) from 0 to 1 with \(a_Q=0.7\), and \(a_Q\) from 0 to 2 with \(\pi_Q=0.2\). To isolate intervention error from sampling error, we compute \(G=\mathbb E[zz^\top]\) analytically. We measure recovery error by \(\|P_2(G)-P_C\|_{\mathrm{op}}\), where \(P_2(G)\) projects onto the two leading eigenvectors of \(G\), and \(P_C=U_CU_C^\top\) projects onto the true nuisance subspace.

\begin{figure}[!t]
\centering
\includegraphics[width=\textwidth]{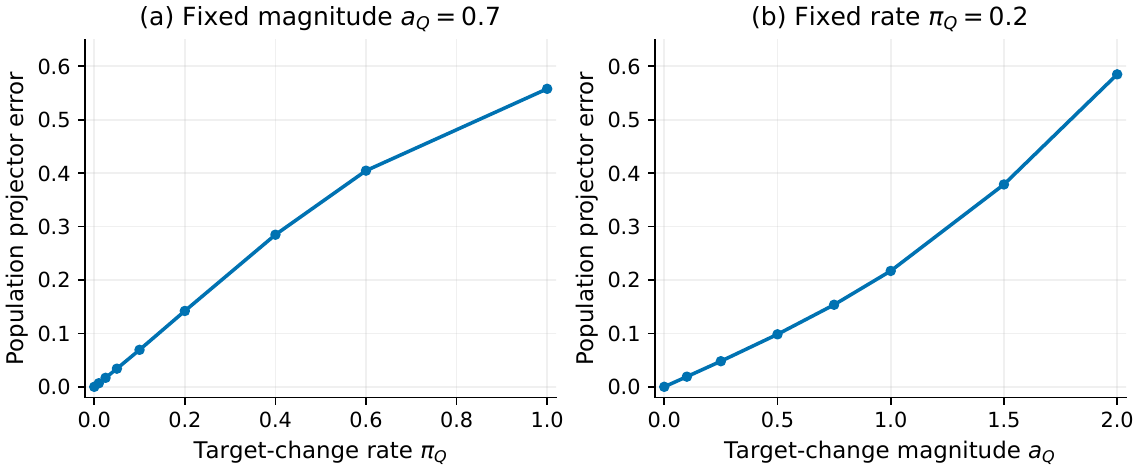}
\caption{Nuisance-subspace recovery deteriorates as unintended target changes become more frequent (left) or larger (right). Results use exact population moments, without sampling error.}
\label{fig:synthetic-error-floor}
\end{figure}

\paragraph{Results.}
\Cref{fig:synthetic-error-floor} shows that nuisance-subspace recovery deteriorates as a larger fraction of interventions change the target, or as those changes grow in magnitude. This error persists even with exact population moments, implying collecting more intervention pairs cannot eliminate it.

\subsection{Attenuation Strength and Target Preservation}
\label{app:synthetic-attenuation}

We examine how attenuation strength affects nuisance suppression and target preservation when the estimated nuisance directions are inaccurate.

\paragraph{Setup.}
Let \(x=U_Qq+U_Cc\), with independent standard Gaussian \(q,c\), and define the true attenuation operator as \(T=U_C\operatorname{diag}(1,0.25)U_C^\top\). We introduce estimation error by rotating the estimated nuisance directions toward the target subspace:
\[
    V_\theta=\cos\theta\,U_C+\sin\theta\,U_Q,\qquad
    \widehat T=V_\theta\operatorname{diag}(1,0.25)V_\theta^\top,
\]
with \(\theta\in\{0,5,15,30\}\) degrees. The true target and nuisance subspaces remain orthogonal. For \(\alpha\in\{0,0.25,0.5,0.75,1\}\), we measure
\[
\underbrace{\|(I-\alpha\widehat T)U_C\|_{\mathrm{op}}}
_{\text{remaining nuisance sensitivity}},
\qquad
\underbrace{\|\alpha\widehat T U_Q\|_{\mathrm{op}}}
_{\text{target distortion}}.
\]
These quantities measure the largest remaining nuisance response and the largest change to the target component over unit-norm factor vectors. We quantify operator error by \(\eta=\|\widehat T-T\|_{\mathrm{op}}\).

\begin{figure}[!t]
\centering
\includegraphics[width=\textwidth]{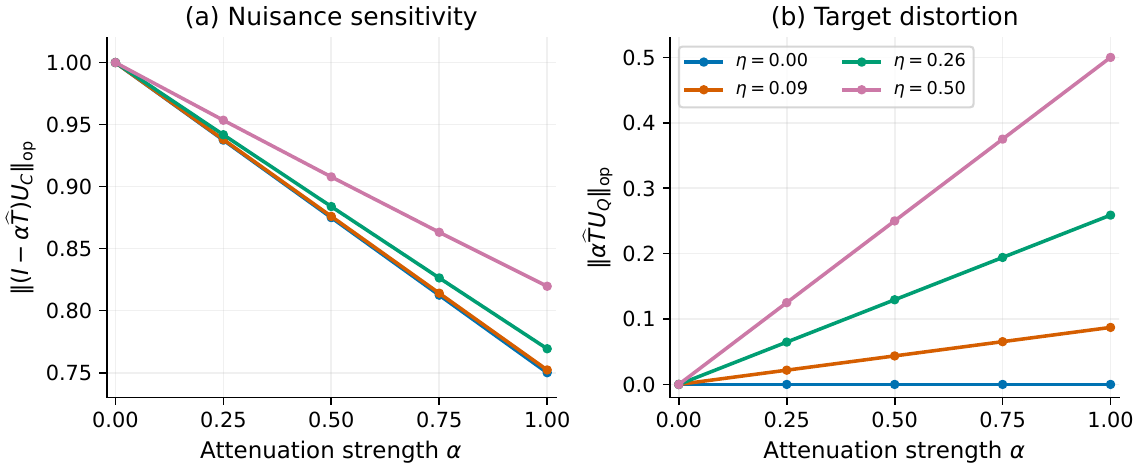}
\caption{Increasing attenuation strength reduces nuisance sensitivity (left) but distorts target information when the estimated nuisance directions are inaccurate (right). Colors indicate operator error \(\eta\); both metrics are computed analytically.}
\label{fig:synthetic-attenuation}
\end{figure}

\paragraph{Results.}
\Cref{fig:synthetic-attenuation} shows that stronger attenuation reduces nuisance sensitivity while preserving the target when the operator is exact. With estimation error, stronger attenuation also increases target distortion, illustrating the tradeoff characterized in \cref{thm:attenuation}. Even with the exact operator and \(\alpha=1\), nuisance sensitivity remains at \(0.75\): attenuation removes the strongest nuisance direction but only partially suppresses the weaker one.

\subsection{Effect of Centering on Nuisance Recovery}
\label{app:synthetic-mean-shift}

\voirdire{} constructs its calibration matrix from paired score shifts without subtracting their mean (\cref{eq:family-shift-covariance}). Here, we illustrate why: a consistent shift across pairs can reveal nuisance sensitivity, and centering would remove this signal.

\paragraph{Setup.}
We model paired shifts as \(z=U_C(g,a)^\top+\xi\), with independent \(g\sim N(0,1)\) and \(\xi\sim N(0,0.0025I_p)\). The first nuisance component varies across pairs, while the second produces a constant shift of magnitude \(a\). Using \(m=128\) pairs, we compare the uncentered second moment \(\widehat G=m^{-1}\sum_i z_i z_i^\top\) with the centered covariance \(\widehat C=m^{-1}\sum_i(z_i-\bar z)(z_i-\bar z)^\top\). We vary \(a\) from 0 to 4 and measure nuisance-subspace recovery using \(\|P_2(\cdot)-P_C\|_{\mathrm{op}}\).

\begin{figure}[!t]
\centering
\includegraphics[width=0.62\textwidth]{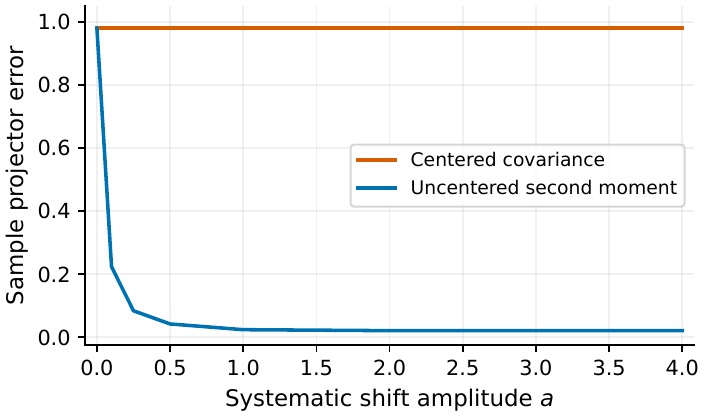}
\caption{Centering discards the systematic shift that identifies the second nuisance direction. The uncentered estimator recovers this direction as the shift grows. At \(a=0\), neither population moment contains information about the second direction.}
\label{fig:synthetic-mean-shift}
\end{figure}

\paragraph{Results.}
\Cref{fig:synthetic-mean-shift} shows that the centered estimator fails to recover the second nuisance direction, while the uncentered estimator becomes more accurate as the systematic shift grows. This supports retaining the mean shift: a consistent response across pairs can reveal nuisance sensitivity even without pair-to-pair variation.

\subsection{Nuisance Recovery and Aggregation Performance}
\label{app:synthetic-clean-recovery}

We examine how nuisance-subspace recovery relates to aggregation performance as the calibration budget increases.

\paragraph{Setup.}
We generate judge outputs as \(x_i=U_Qq_i+U_Cc_i+\epsilon_i\), with independent target factors \(q_i\sim N(0,\operatorname{diag}(1,0.7^2))\), nuisance factors \(c_i\sim N(0,\operatorname{diag}(3^2,2.5^2))\), and Gaussian noise \(\epsilon_i\) correlated across judges, with scale \(0.5\). The target score is \(y_i=0.8q_{i1}+0.6q_{i2}+\nu_i\), where \(\nu_i\sim N(0,0.1^2)\) is independent. We use 5,000 training and 2,000 test items. Two intervention families each add an \(N(2,0.4^2)\) shift to a different nuisance coordinate, preserving \(q_i\) and resampling judge noise. Each family supplies \(m\in\{4,8,16,32,64,128,256,512\}\) pairs, for \(2m\) pairs in total.

\voirdire{} estimates aggregation weights from all training outputs and selects \((\alpha,\gamma)\) separately for each seed and budget using MAE on a 20\% validation subset and the grids in \cref{app:current-protocol}. We compare against PCA, CARE-SVD, and an oracle reference that projects out the true nuisance subspace using \(I-P_C\) before fitting a two-component PCA aggregator. The oracle orients each eigenvector to have a positive coordinate sum, combines them with weights proportional to the square roots of their eigenvalues, and normalizes the resulting vector to sum to one. It knows the nuisance subspace but not the target coefficients. We report test MAE and nuisance-subspace recovery error \(\|P_2(\widehat G)-P_C\|_{\mathrm{op}}\).

\begin{figure}[!t]
\centering
\includegraphics[width=\textwidth]{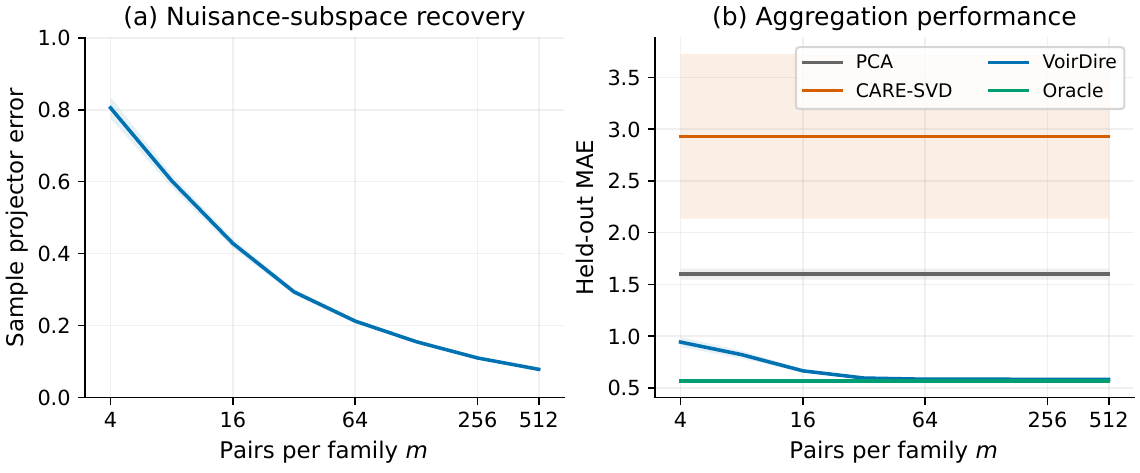}
\caption{Increasing the calibration budget improves nuisance-subspace recovery (left) and aggregation performance (right). The oracle reference removes the true nuisance subspace before PCA aggregation. Budgets are per intervention family.}
\label{fig:synthetic-target-preserving-mechanism}
\end{figure}

\paragraph{Results.}
\Cref{fig:synthetic-target-preserving-mechanism} shows that better nuisance recovery accompanies lower aggregation error. As the budget increases from 4 to 512 pairs per family, \voirdire{}'s MAE falls from \(0.942\) to \(0.582\), approaching the oracle reference's \(0.567\). Most of the aggregation gain occurs by 32--64 pairs per family, although subspace recovery continues to improve at larger budgets. These results show that \voirdire{} can improve aggregation even when the nuisance subspace is estimated imperfectly.
\section{Additional Experimental Results}
\label{app:full-experiment-results}

We provide further analyses of attenuation strength and judge-panel size (\cref{app:attenuation-strength-ablation,app:judge-panel-size-ablation}). We also report calibration-budget results for targeted nuisance removal (\cref{app:injected-bias-results}) and compare the effects of score calibration across aggregators (\cref{app:gain-decomposition}).

\subsection{Sensitivity to Attenuation and Regularization}
\label{app:attenuation-strength-ablation}

We study how attenuation strength \(\alpha\) and regularization multiplier \(\gamma\) affect aggregation performance.

\paragraph{Setup.}
We use the datasets, splits, calibration pairs, and selected intervention families from \cref{subsec:exp_main_agg}. We evaluate all 30 combinations of \(\alpha\in\{0,0.1,0.25,0.5,0.75,1\}\) and \(\gamma\in\{0.25,1,4,16,64\}\), reporting MAE for continuous tasks and accuracy for binary/preference tasks. \Cref{tab:gamma-sensitivity} reports sensitivity to \(\gamma\) at the validation-selected \(\alpha\); \cref{tab:loo} reports the effect of removing each selected intervention family and refitting.

\begin{figure*}[!t]
\centering
\includegraphics[width=\textwidth]{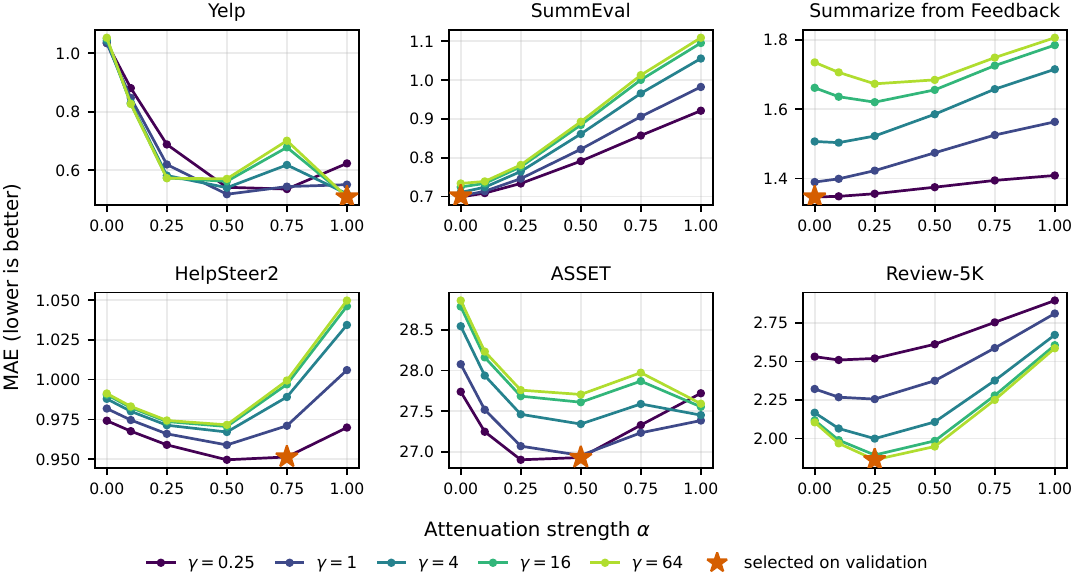}
\caption{Sensitivity to \(\alpha\) and \(\gamma\) on continuous-score datasets. Curves show mean evaluation MAE across five splits for each \(\gamma\); lower is better. Stars mark validation-selected configurations.}
\label{fig:alpha-gamma-continuous}
\end{figure*}

\begin{figure*}[!t]
\centering
\includegraphics[width=\textwidth]{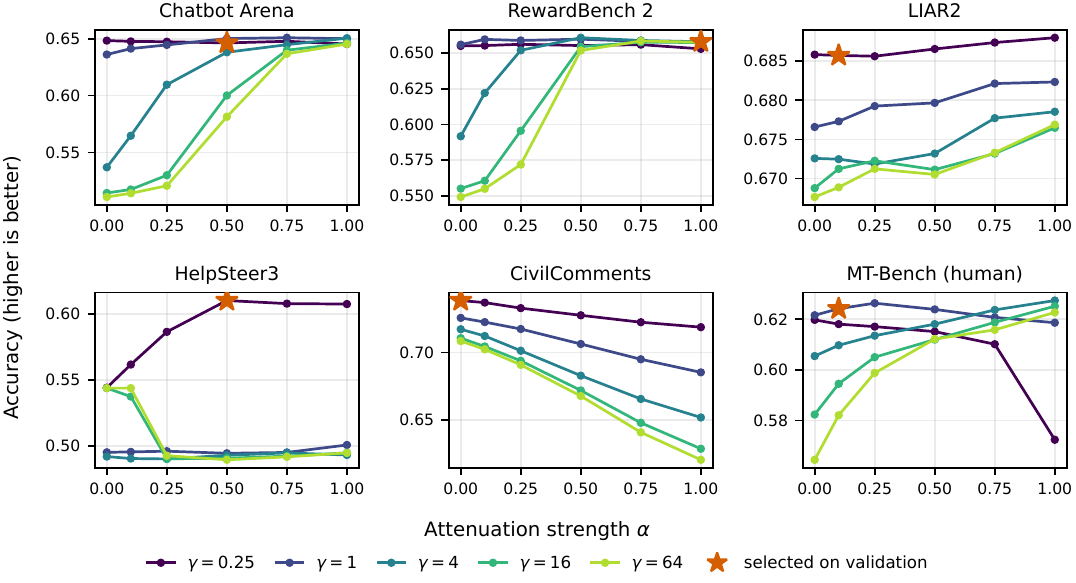}
\caption{Sensitivity to \(\alpha\) and \(\gamma\) on binary/preference datasets. Curves show mean evaluation accuracy across five splits for each \(\gamma\); higher is better. Stars mark validation-selected configurations.}
\label{fig:alpha-gamma-binary}
\end{figure*}

\begin{table}[!t]
    \centering
    \footnotesize
    \setlength{\tabcolsep}{2.5pt}
    \begin{tabular}{lcccccc}
        \toprule
        Dataset & $\alpha$ & $\gamma{=}0.25$ & $1$ & $4$ & $16$ & $64$ \\
        \midrule
        \multicolumn{7}{l}{\emph{Continuous (MAE, lower is better)}} \\
        Yelp & 1 & 0.623 & 0.550 & 0.519 & 0.510 & \textbf{0.508} \\
        SummEval & 0 & 0.699 & \textbf{0.701} & 0.710 & 0.724 & 0.733 \\
        Summarize from Feedback & 0 & \textbf{1.346} & 1.389 & 1.507 & 1.662 & 1.735 \\
        HelpSteer2 & 0.75 & \textbf{0.951} & 0.971 & 0.989 & 0.997 & 0.999 \\
        ASSET & 0.5 & \textbf{26.930} & 26.956 & 27.342 & 27.611 & 27.706 \\
        Review-5K & 0.25 & 2.519 & 2.256 & 2.000 & 1.893 & \textbf{1.862} \\
        \midrule
        \multicolumn{7}{l}{\emph{Binary/preference (accuracy, higher is better)}} \\
        Chatbot Arena & 0.5 & \textbf{0.646} & 0.650 & 0.638 & 0.600 & 0.581 \\
        RewardBench 2 & 1 & 0.653 & 0.658 & \textbf{0.658} & 0.657 & 0.658 \\
        LIAR2 & 0.1 & \textbf{0.686} & 0.677 & 0.672 & 0.671 & 0.669 \\
        HelpSteer3 & 0.5 & \textbf{0.610} & 0.495 & 0.493 & 0.491 & 0.490 \\
        CivilComments & 0 & \textbf{0.739} & 0.726 & 0.717 & 0.711 & 0.709 \\
        MT-Bench (human) & 0.1 & 0.618 & \textbf{0.624} & 0.610 & 0.595 & 0.582 \\
        \bottomrule
    \end{tabular}
    \caption{Sensitivity to the regularization multiplier $\gamma$ at the selected $\alpha$ and family set of the main tables. Entries are mean evaluation metrics over the protocol seeds; the value selected on validation is in bold.}
    \label{tab:gamma-sensitivity}
\end{table}

\begin{table}[!t]
    \centering
    \footnotesize
    \setlength{\tabcolsep}{2.5pt}
    \begin{tabular}{lccccc}
        \toprule
        Dataset & $k$ & Help & Mean $\Delta$ & Max loss & Max gain \\
        \midrule
        \multicolumn{6}{l}{\emph{Continuous ($\Delta$ = MAE without the family $-$ MAE with it)}} \\
        Yelp & 4 & 4 & 0.024 & 0.072 & 0.003 \\
        SummEval & 4 & 3 & 0.001 & 0.002 & 0.000 \\
        Summarize from Feedback & 4 & 4 & 0.009 & 0.024 & 0.001 \\
        HelpSteer2 & 3 & 3 & 0.014 & 0.023 & 0.005 \\
        ASSET & 5 & 2 & 0.024 & 0.263 & -0.132 \\
        Review-5K & 4 & 3 & 0.021 & 0.063 & -0.009 \\
        \midrule
        \multicolumn{6}{l}{\emph{Binary/preference ($\Delta$ = accuracy without $-$ accuracy with)}} \\
        Chatbot Arena & 5 & 5 & -0.013 & -0.038 & -0.003 \\
        RewardBench 2 & 4 & 2 & -0.036 & -0.151 & 0.007 \\
        LIAR2 & 2 & 2 & -0.002 & -0.003 & -0.001 \\
        HelpSteer3 & 6 & 6 & -0.026 & -0.070 & -0.004 \\
        CivilComments & 2 & 1 & -0.006 & -0.020 & 0.007 \\
        MT-Bench (human) & 4 & 3 & -0.032 & -0.124 & 0.001 \\
        \bottomrule
    \end{tabular}
    \caption{Leave-one-family-out refits of the selected family sets, $(\alpha,\gamma)$ re-selected without the dropped family. $k$: number of selected families. Help: families whose removal worsens the evaluation metric. Mean $\Delta$: mean change over the $k$ refits. Max loss and max gain: the largest deterioration and the largest improvement obtained by dropping one family. Means over the protocol seeds.}
    \label{tab:loo}
\end{table}

\paragraph{Results.}
\Cref{fig:alpha-gamma-continuous,fig:alpha-gamma-binary} show that the benefit of attenuation varies across datasets. Yelp benefits from strong attenuation, whereas Review-5K and ASSET deteriorate beyond moderate attenuation. On SummEval, Summarize from Feedback, and CivilComments, validation selects \(\alpha=0\), retaining the original scores while using interventions to choose aggregation weights.

The effect of \(\alpha\) also depends on \(\gamma\). On Chatbot Arena, attenuation provides larger gains at \(\gamma=64\), where aggregation places less emphasis on intervention sensitivity. At \(\gamma=0.25\), varying \(\alpha\) changes accuracy by less than one percentage point. These results support tuning attenuation and regularization jointly.

\subsection{Judge-Panel Size Ablation}
\label{app:judge-panel-size-ablation}
We examine how the number of judges affects aggregation performance and the gains from intervention-based calibration.

\paragraph{Setup.}
We use the 12 datasets from \cref{subsec:exp_main_agg}, varying \(K\in\{2,4,6,8,11\}\) for continuous tasks and \(K\in\{2,4,8,12,16,20\}\) for binary/preference tasks. Twenty random orderings of the judges define nested subsets shared across methods. We keep the data splits, intervention families, and calibration pairs fixed, and refit and tune each method for each subset. We average performance over five splits and then across distinct subsets; bands show one standard deviation across subset means.

\begin{figure*}[!t]
    \centering
    \includegraphics[width=\textwidth]{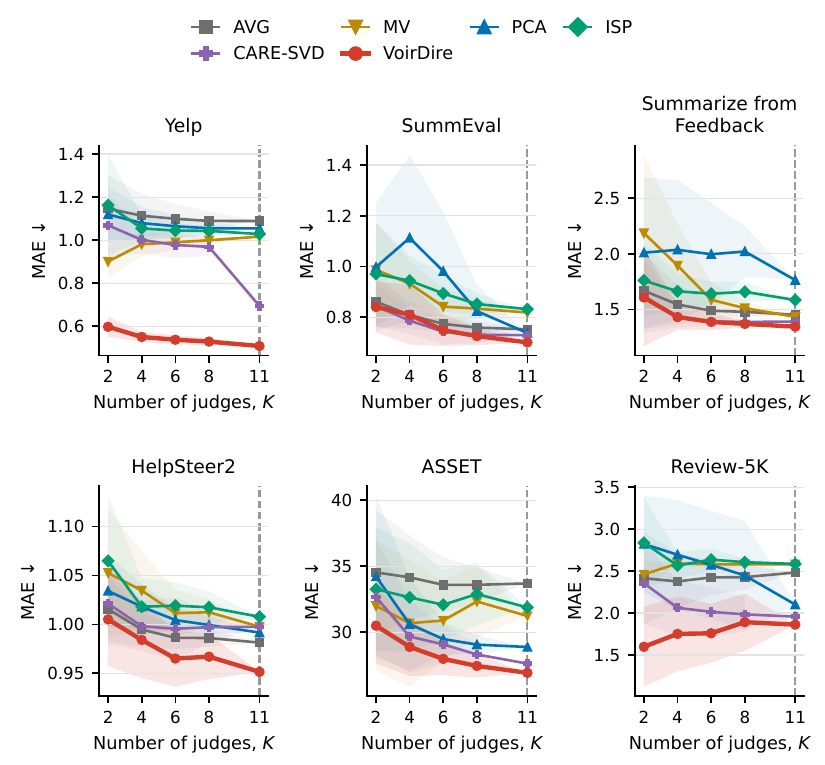}
    \caption{MAE versus number of judges on continuous-score tasks; lower is better. Bands show $\pm1$ standard deviation across subset means. Vertical dashed lines mark the main panel size.}
    \label{fig:panel-size-continuous}
    \vspace{-5mm}
\end{figure*}

\begin{figure*}[!t]
    \centering
    \includegraphics[width=\textwidth]{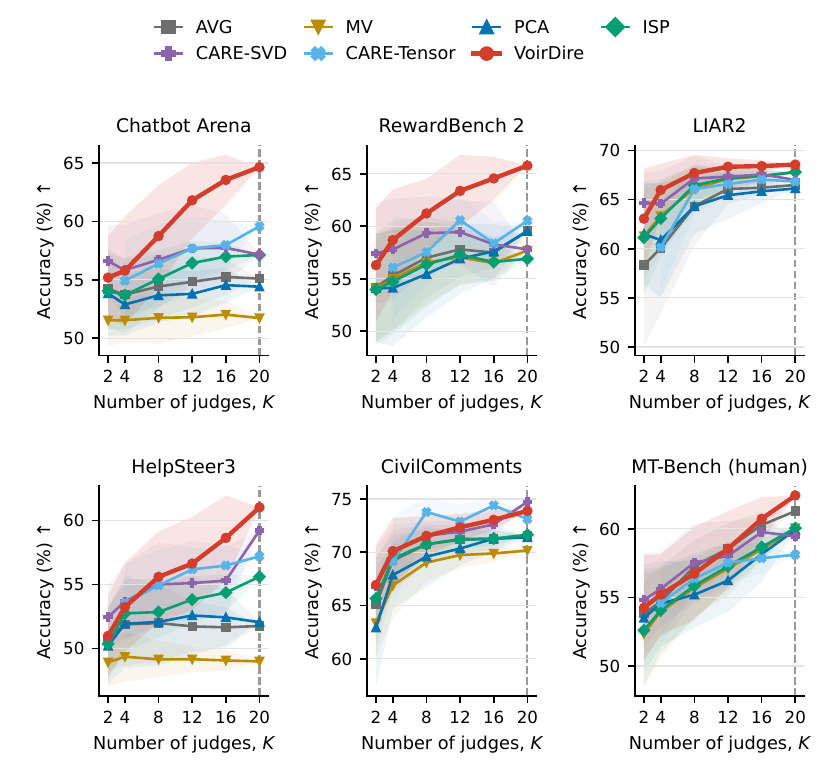}
    \caption{Accuracy versus number of judges on binary/preference tasks; higher is better. Bands show $\pm1$ standard deviation across subset means. The vertical dashed line marks the main panel size. CARE-Tensor is unavailable at \(K=2\).}
    \label{fig:panel-size-binary}
    \vspace{-5mm}
\end{figure*}

\paragraph{Results.}
\Cref{fig:panel-size-continuous,fig:panel-size-binary} show that larger judge panels generally improve aggregation. From the smallest to the full panel, \voirdire{} improves on five of six continuous tasks and all six binary/preference tasks, with accuracy gains of 5.51--10.05 percentage points. The exception is Review-5K, where MAE rises from \(1.598\) to \(1.862\). Overall, larger panels tend to improve aggregation performance, although the benefits vary across methods and datasets.

\subsection{Calibration Budget for Targeted Nuisance Removal}
\label{app:injected-bias-results}

We examine how the number of calibration pairs affects sensitivity to the beauty and authority biases studied in \cref{subsec:injected-bias}.

\paragraph{Setup.}
Following the protocol in \cref{app:injected-bias}, we vary the budget from 4 to 64 pairs using nested calibration sets within each training fold. We keep the evaluation folds and \((\alpha,\gamma)=(1,1/4)\) fixed. We report mean absolute score shifts, averaging first over five folds and then over ten seeds. Variability is measured by the standard deviation across seed means. The main comparison uses the 32-pair setting.

\begin{figure}[t]
\centering
\includegraphics[width=\linewidth]{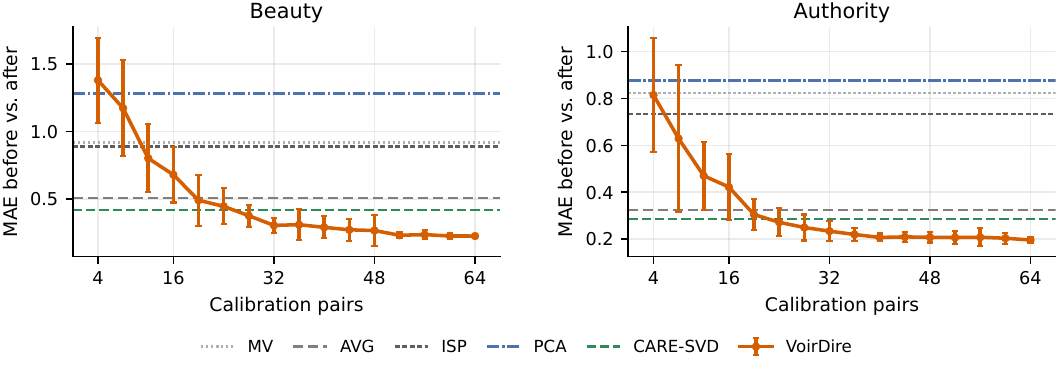}
\caption{Sensitivity to injected biases versus calibration pairs, with $(\alpha,\gamma)=(1,1/4)$. Error bars show standard deviation across ten seed means. Baselines use no calibration pairs. Lower is better.}
\label{fig:injected-bias-budget}
\vspace{-5mm}
\end{figure}

\paragraph{Results.}
\Cref{fig:injected-bias-budget} shows that increasing the number of calibration pairs improves suppression of the targeted nuisance. This is consistent with \cref{thm:voirdire-sample}, which bounds the sampling error in the calibration matrix as the pair budget grows. With 32 pairs, \voirdire{} achieves lower sensitivity than all baselines on both beauty and authority (\cref{tab:injected-bias-m32}).

\subsection{Effect of Calibration on Other Aggregators}
\label{app:gain-decomposition}

We extend \cref{subsec:gain-decomposition} to examine whether score calibration benefits other aggregation methods.

\paragraph{Setup.}
We follow \cref{subsec:exp_main_agg}, applying each aggregator to both raw and calibrated judge scores with the same selected interventions and calibration settings. MV and ISP receive calibrated scores mapped to the original score levels. The intervention-aware aggregation conditions follow \cref{subsec:gain-decomposition}. When the selected \(\alpha=0\), raw and calibrated scores coincide.


\begin{table*}[!t]
    \centering
    \small
    \setlength{\tabcolsep}{3.0pt}
    \resizebox{\textwidth}{!}{%
    \begin{tabular}{llcccccc}
        \toprule
        & & \multicolumn{6}{c}{Aggregator} \\
        \cmidrule(lr){3-8}
        Dataset & Condition & MV & AVG & ISP & CARE-SVD & PCA & \shortstack{Intervention-\\aware} \\
        \midrule
        Yelp & Raw & $1.016\pm0.002$ & $1.089\pm0.002$ & $1.028\pm0.002$ & $0.694\pm0.004$ & $1.054\pm0.002$ & $1.034\pm0.007$ \\
         & Calibrated & $1.398\pm0.006$ & $1.605\pm0.112$ & $1.347\pm0.090$ & $7.098\pm8.498$ & $0.507\pm0.016$ & $0.508\pm0.016$ \\
         & $\Delta$ & \textcolor{red!70!black}{$-0.381$} & \textcolor{red!70!black}{$-0.516$} & \textcolor{red!70!black}{$-0.318$} & \textcolor{red!70!black}{$-6.404$} & \textcolor{green!45!black}{$+0.547$} & \textcolor{green!45!black}{$+0.526$} \\
        \addlinespace[1pt]
        \multicolumn{8}{@{}c@{}}{\leavevmode\color{black!35}\leaders\hbox{\rule{0.8em}{0.35pt}\hskip0.5em}\hfill\kern0pt}\\[-0.4ex]
        SummEval & Raw & $0.819\pm0.005$ & $0.753\pm0.004$ & $0.832\pm0.005$ & $0.729\pm0.005$ & $0.738\pm0.004$ & $0.701\pm0.005$ \\
         & Calibrated & $0.819\pm0.005$ & $0.753\pm0.004$ & $0.832\pm0.005$ & $0.729\pm0.005$ & $0.738\pm0.004$ & $0.701\pm0.005$ \\
         & $\Delta$ & \textcolor{black!60}{$+0.000$} & \textcolor{black!60}{$+0.000$} & \textcolor{black!60}{$+0.000$} & \textcolor{black!60}{$+0.000$} & \textcolor{black!60}{$+0.000$} & \textcolor{black!60}{$+0.000$} \\
        \addlinespace[1pt]
        \multicolumn{8}{@{}c@{}}{\leavevmode\color{black!35}\leaders\hbox{\rule{0.8em}{0.35pt}\hskip0.5em}\hfill\kern0pt}\\[-0.4ex]
        Summarize from Feedback & Raw & $1.434\pm0.004$ & $1.453\pm0.003$ & $1.585\pm0.003$ & $1.392\pm0.014$ & $1.764\pm0.004$ & $1.346\pm0.004$ \\
         & Calibrated & $1.434\pm0.004$ & $1.453\pm0.003$ & $1.585\pm0.003$ & $1.392\pm0.014$ & $1.764\pm0.004$ & $1.346\pm0.004$ \\
         & $\Delta$ & \textcolor{black!60}{$+0.000$} & \textcolor{black!60}{$+0.000$} & \textcolor{black!60}{$+0.000$} & \textcolor{black!60}{$+0.000$} & \textcolor{black!60}{$+0.000$} & \textcolor{black!60}{$+0.000$} \\
        \addlinespace[1pt]
        \multicolumn{8}{@{}c@{}}{\leavevmode\color{black!35}\leaders\hbox{\rule{0.8em}{0.35pt}\hskip0.5em}\hfill\kern0pt}\\[-0.4ex]
        HelpSteer2 & Raw & $0.998\pm0.011$ & $0.981\pm0.006$ & $1.008\pm0.012$ & $0.998\pm0.012$ & $0.992\pm0.006$ & $0.974\pm0.005$ \\
         & Calibrated & $1.006\pm0.059$ & $1.039\pm0.087$ & $0.997\pm0.037$ & $0.968\pm0.017$ & $1.000\pm0.048$ & $0.951\pm0.013$ \\
         & $\Delta$ & \textcolor{red!70!black}{$-0.008$} & \textcolor{red!70!black}{$-0.057$} & \textcolor{green!45!black}{$+0.010$} & \textcolor{green!45!black}{$+0.030$} & \textcolor{red!70!black}{$-0.009$} & \textcolor{green!45!black}{$+0.023$} \\
        \addlinespace[1pt]
        \multicolumn{8}{@{}c@{}}{\leavevmode\color{black!35}\leaders\hbox{\rule{0.8em}{0.35pt}\hskip0.5em}\hfill\kern0pt}\\[-0.4ex]
        ASSET & Raw & $31.238\pm0.118$ & $33.701\pm0.074$ & $31.885\pm0.108$ & $27.629\pm0.156$ & $28.888\pm0.132$ & $27.740\pm0.472$ \\
         & Calibrated & $31.658\pm1.704$ & $31.872\pm0.148$ & $29.723\pm1.725$ & $27.804\pm0.745$ & $27.741\pm0.177$ & $26.930\pm0.342$ \\
         & $\Delta$ & \textcolor{red!70!black}{$-0.420$} & \textcolor{green!45!black}{$+1.829$} & \textcolor{green!45!black}{$+2.162$} & \textcolor{red!70!black}{$-0.175$} & \textcolor{green!45!black}{$+1.147$} & \textcolor{green!45!black}{$+0.809$} \\
        \addlinespace[1pt]
        \multicolumn{8}{@{}c@{}}{\leavevmode\color{black!35}\leaders\hbox{\rule{0.8em}{0.35pt}\hskip0.5em}\hfill\kern0pt}\\[-0.4ex]
        Review-5K & Raw & $2.581\pm0.000$ & $2.483\pm0.000$ & $2.585\pm0.000$ & $1.957\pm0.018$ & $2.101\pm0.000$ & $2.105\pm0.001$ \\
         & Calibrated & $2.408\pm0.163$ & $1.797\pm0.076$ & $2.339\pm0.180$ & $1.475\pm0.148$ & $1.852\pm0.033$ & $1.862\pm0.035$ \\
         & $\Delta$ & \textcolor{green!45!black}{$+0.174$} & \textcolor{green!45!black}{$+0.686$} & \textcolor{green!45!black}{$+0.246$} & \textcolor{green!45!black}{$+0.482$} & \textcolor{green!45!black}{$+0.249$} & \textcolor{green!45!black}{$+0.243$} \\
        \bottomrule
    \end{tabular}
    }
    \caption{Effect of score calibration on continuous-score aggregation.
    Entries show mean $\pm$ standard deviation of MAE across five
    splits; lower is better.
    \(\Delta=\text{Raw}-\text{Calibrated}\), so positive values
    indicate lower calibrated MAE.
    The calibrated intervention-aware aggregator is \voirdire{};
    its raw counterpart retains the intervention penalty.
    Fitting details appear in \cref{app:gain-decomposition}.}
    \label{tab:appendix-final-aggregator-continuous}
\end{table*}

\begin{table*}[!t]
    \centering
    \small
    \setlength{\tabcolsep}{3.0pt}
    \resizebox{\textwidth}{!}{%
    \begin{tabular}{llccccccc}
        \toprule
        & & \multicolumn{7}{c}{Aggregator} \\
        \cmidrule(lr){3-9}
        Dataset & Condition & MV & AVG & ISP & CARE-SVD & CARE-Tensor & PCA & \shortstack{Intervention-\\aware} \\
        \midrule
        Chatbot Arena & Raw & $0.517\pm0.003$ & $0.551\pm0.004$ & $0.571\pm0.003$ & $0.572\pm0.016$ & $0.564\pm0.028$ & $0.544\pm0.002$ & $0.649\pm0.006$ \\
         & Calibrated & $0.517\pm0.003$ & $0.570\pm0.003$ & $0.577\pm0.002$ & $0.612\pm0.021$ & $0.549\pm0.032$ & $0.534\pm0.026$ & $0.646\pm0.011$ \\
         & $\Delta$ & \textcolor{black!60}{$+0.000$} & \textcolor{green!45!black}{$+0.019$} & \textcolor{green!45!black}{$+0.006$} & \textcolor{green!45!black}{$+0.040$} & \textcolor{red!70!black}{$-0.015$} & \textcolor{red!70!black}{$-0.010$} & \textcolor{red!70!black}{$-0.002$} \\
        \addlinespace[1pt]
        \multicolumn{9}{@{}c@{}}{\leavevmode\color{black!35}\leaders\hbox{\rule{0.8em}{0.35pt}\hskip0.5em}\hfill\kern0pt}\\[-0.4ex]
        RewardBench 2 & Raw & $0.577\pm0.003$ & $0.596\pm0.003$ & $0.569\pm0.003$ & $0.578\pm0.015$ & $0.545\pm0.036$ & $0.595\pm0.004$ & $0.655\pm0.013$ \\
         & Calibrated & $0.565\pm0.006$ & $0.597\pm0.005$ & $0.590\pm0.006$ & $0.561\pm0.066$ & $0.618\pm0.019$ & $0.547\pm0.011$ & $0.658\pm0.006$ \\
         & $\Delta$ & \textcolor{red!70!black}{$-0.012$} & \textcolor{green!45!black}{$+0.001$} & \textcolor{green!45!black}{$+0.021$} & \textcolor{red!70!black}{$-0.017$} & \textcolor{green!45!black}{$+0.073$} & \textcolor{red!70!black}{$-0.048$} & \textcolor{green!45!black}{$+0.003$} \\
        \addlinespace[1pt]
        \multicolumn{9}{@{}c@{}}{\leavevmode\color{black!35}\leaders\hbox{\rule{0.8em}{0.35pt}\hskip0.5em}\hfill\kern0pt}\\[-0.4ex]
        LIAR2 & Raw & $0.678\pm0.004$ & $0.665\pm0.004$ & $0.678\pm0.004$ & $0.670\pm0.007$ & $0.659\pm0.029$ & $0.661\pm0.004$ & $0.686\pm0.004$ \\
         & Calibrated & $0.678\pm0.004$ & $0.669\pm0.004$ & $0.678\pm0.004$ & $0.684\pm0.014$ & $0.672\pm0.013$ & $0.679\pm0.004$ & $0.686\pm0.005$ \\
         & $\Delta$ & \textcolor{black!60}{$+0.000$} & \textcolor{green!45!black}{$+0.005$} & \textcolor{black!60}{$+0.000$} & \textcolor{green!45!black}{$+0.014$} & \textcolor{green!45!black}{$+0.014$} & \textcolor{green!45!black}{$+0.018$} & \textcolor{red!70!black}{$-0.000$} \\
        \addlinespace[1pt]
        \multicolumn{9}{@{}c@{}}{\leavevmode\color{black!35}\leaders\hbox{\rule{0.8em}{0.35pt}\hskip0.5em}\hfill\kern0pt}\\[-0.4ex]
        HelpSteer3 & Raw & $0.490\pm0.001$ & $0.517\pm0.002$ & $0.556\pm0.007$ & $0.592\pm0.027$ & $0.567\pm0.009$ & $0.520\pm0.004$ & $0.544\pm0.003$ \\
         & Calibrated & $0.490\pm0.001$ & $0.515\pm0.004$ & $0.556\pm0.007$ & $0.551\pm0.024$ & $0.562\pm0.006$ & $0.530\pm0.017$ & $0.610\pm0.015$ \\
         & $\Delta$ & \textcolor{black!60}{$+0.000$} & \textcolor{red!70!black}{$-0.002$} & \textcolor{black!60}{$+0.000$} & \textcolor{red!70!black}{$-0.041$} & \textcolor{red!70!black}{$-0.004$} & \textcolor{green!45!black}{$+0.010$} & \textcolor{green!45!black}{$+0.066$} \\
        \addlinespace[1pt]
        \multicolumn{9}{@{}c@{}}{\leavevmode\color{black!35}\leaders\hbox{\rule{0.8em}{0.35pt}\hskip0.5em}\hfill\kern0pt}\\[-0.4ex]
        CivilComments & Raw & $0.702\pm0.001$ & $0.715\pm0.002$ & $0.716\pm0.002$ & $0.747\pm0.003$ & $0.735\pm0.011$ & $0.714\pm0.001$ & $0.739\pm0.003$ \\
         & Calibrated & $0.702\pm0.001$ & $0.715\pm0.002$ & $0.716\pm0.002$ & $0.747\pm0.003$ & $0.735\pm0.011$ & $0.714\pm0.001$ & $0.739\pm0.003$ \\
         & $\Delta$ & \textcolor{black!60}{$+0.000$} & \textcolor{black!60}{$+0.000$} & \textcolor{black!60}{$+0.000$} & \textcolor{black!60}{$+0.000$} & \textcolor{black!60}{$+0.000$} & \textcolor{black!60}{$+0.000$} & \textcolor{black!60}{$+0.000$} \\
        \addlinespace[1pt]
        \multicolumn{9}{@{}c@{}}{\leavevmode\color{black!35}\leaders\hbox{\rule{0.8em}{0.35pt}\hskip0.5em}\hfill\kern0pt}\\[-0.4ex]
        MT-Bench (human) & Raw & $0.600\pm0.004$ & $0.613\pm0.004$ & $0.600\pm0.004$ & $0.594\pm0.005$ & $0.550\pm0.017$ & $0.599\pm0.004$ & $0.622\pm0.010$ \\
         & Calibrated & $0.600\pm0.004$ & $0.612\pm0.004$ & $0.600\pm0.004$ & $0.603\pm0.019$ & $0.548\pm0.033$ & $0.601\pm0.004$ & $0.624\pm0.007$ \\
         & $\Delta$ & \textcolor{black!60}{$+0.000$} & \textcolor{red!70!black}{$-0.001$} & \textcolor{black!60}{$+0.000$} & \textcolor{green!45!black}{$+0.009$} & \textcolor{red!70!black}{$-0.002$} & \textcolor{green!45!black}{$+0.002$} & \textcolor{green!45!black}{$+0.003$} \\
        \bottomrule
    \end{tabular}
    }
    \caption{Effect of score calibration on binary/preference aggregation.
    Entries show mean $\pm$ standard deviation of accuracy across
    five splits; higher is better.
    \(\Delta=\text{Calibrated}-\text{Raw}\), so positive values
    indicate improvement. Accuracy and differences are reported
    as proportions.
    The calibrated intervention-aware aggregator is \voirdire{};
    its raw counterpart retains the intervention penalty.
    Fitting details appear in \cref{app:gain-decomposition}.}
    \label{tab:appendix-final-aggregator-binary}
\end{table*}

\paragraph{Results.}
\Cref{tab:appendix-final-aggregator-continuous,tab:appendix-final-aggregator-binary} show that calibration benefits depend on both the dataset and the aggregation rule. Every aggregator improves on Review-5K, whereas on Yelp, calibration improves PCA and intervention-aware aggregation but worsens MV, AVG, ISP, and CARE-SVD. For intervention-aware aggregation, calibration lowers MAE on all four continuous datasets with \(\alpha>0\) and improves accuracy on three of the five corresponding binary/preference datasets. Thus, score calibration can benefit other aggregators, but is not a uniformly beneficial preprocessing step.

\end{document}